\pdfoutput=1

\RequirePackage{fix-cm}
\documentclass[twocolumn]{svjour3}          
\smartqed  
\usepackage{graphicx}

\usepackage{amsmath} 
\usepackage{amssymb}  
\usepackage{color}
\usepackage{url}
\usepackage{multicol}
\usepackage{xcolor,colortbl}
\usepackage{wrapfig}
\usepackage{mdwtab} 
\usepackage{mdwmath}
\usepackage{enumerate}
\usepackage{algorithm2e}
\usepackage[utf8]{inputenc}

\makeatletter
\newcounter{algorithmbis}
\renewcommand{\thealgorithmbis}{\thesection.\arabic{algorithmbis}}
\def\algorithmbis{\@ifnextchar[{\@algorithmbisa}{\@algorithmbisb}}
\def\@algorithmbisa[#1]{%
  \refstepcounter{algorithmbis}
  \trivlist
  \leftmargin\z@
  \itemindent\z@
  \labelsep\z@
  \item[\parbox{\linewidth}{%
    \hrule
    \hrule
    \noindent\strut\textbf{Algorithm \thealgorithmbis} #1
    \hrule
  }]\hfil\vskip0em%
}
\def\@algorithmbisb{\@algorithmbisa[]}

\makeatother

\usepackage{fixltx2e}

\usepackage{xcolor}

\renewcommand{\sl}[1]{\color{green} { #1 } \normalcolor}

\begin{document}

\title{
Towards engagement models that consider individual factors in HRI: on the relation of extroversion and negative attitude towards robot to gaze and speech during a human-robot assembly task
\thanks
{This work was performed within the project EDHHI of Labex SMART (ANR-11-LABX-65) supported by French state funds managed by the ANR within the Investissements d'Avenir programme under reference  ANR-11-IDEX-0004-02. The work was partially supported by the FP7 EU projects CoDyCo (No. 600716 ICT 2011.2.1 Cognitive Systems and Robotics).}
}


\author{
        Serena Ivaldi \and Sebastien Lefort \and Jan Peters \and Mohamed Chetouani \and Joelle Provasi \and Elisabetta Zibetti
}


\institute{S. Ivaldi \at
              Inria, Villers-l\`es-Nancy, F-54600, France\\
	Loria, CNRS \& Universit\'e de Lorraine, Loria, UMR n. 7503, Vandoeuvre-l\`es-Nancy, F-54500, France\\
		Intelligent Autonomous Systems, TU Darmstadt, Germany\\
              Tel.: +33-03-5495-8508\\
              \email{serena.ivaldi@inria.fr}           
           \and        
           S. Lefort \at          
           LIP6, Paris, France          
           \and          
           J. Peters \at
           Intelligent Autonomous Systems, TU Darmstadt, Germany\\
           Max Planck Institute for Intelligent Systems, Germany         
           \and                   
	M. Chetouani \at
	CNRS \& Sorbonne Universit\'es, UPMC Universit\'e Paris 06, Institut des Syst\`emes Intelligents et de Robotique (ISIR) UMR7222, Paris, France
	\and
	J. Provasi \and E. Zibetti \at
	CHARt-Lutin, Universit\'e Paris 8, Paris, France	
}

\date{Received: date / Accepted: date}

\maketitle

\begin{abstract}

Estimating the engagement is critical for human - robot interaction. Engagement measures typically rely on the dynamics of the social signals exchanged by the partners, especially speech and gaze. However, the dynamics of these signals is likely to be influenced by individual and social factors, such as personality traits, as it is well documented that they critically influence how two humans interact with each other.
Here, we assess the influence of two factors, namely extroversion and negative attitude toward robots, on speech and gaze during a cooperative task, where a human must physically manipulate a robot to assemble an object. 
We evaluate if the score of extroversion and negative attitude towards robots co-variate with the duration and frequency of gaze and speech cues.
The experiments were carried out with the humanoid robot iCub and N=56 adult participants.
We found that the more people are extrovert, the more and longer they tend to talk with the robot; and the more people have a negative attitude towards robots, the less they will look at the robot face and the more they will look at the robot hands where the assembly and the contacts occur.
Our results confirm and provide evidence that the engagement models classically used in human-robot interaction should take into account attitudes and personality traits.

\end{abstract}


\section{Introduction}

\begin{figure*}[t]
\centering
{
\includegraphics[width=0.8\hsize]{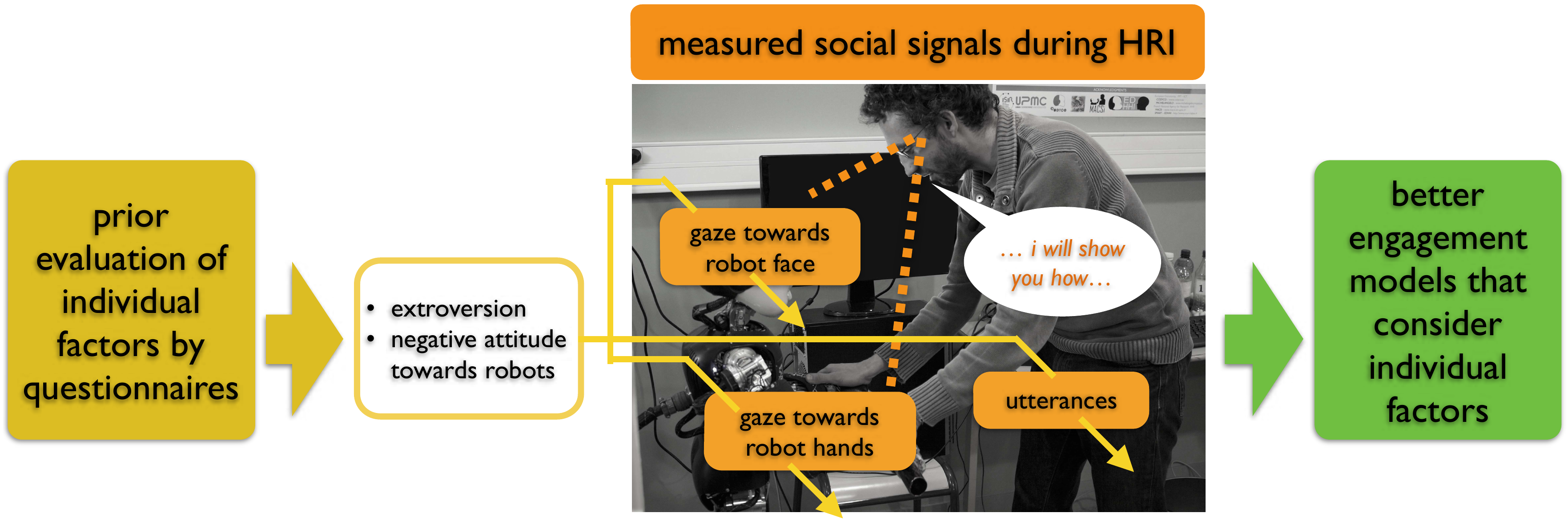}
}
\caption{Conceptual representation of the experiment: we study the relation of extroversion and negative attitude toward robots on speech and gaze during a cooperative assembly task.}
\label{fig:concept}
\end{figure*}

Service and personal robots must be capable of cooperating and interacting with humans for a variety of tasks.
The robot's social skills are crucial to prevent the interaction to become cumbersome and the cooperation less effective.  
Social signals, i.e., verbal and non-verbal cues produced by the human and directed towards the robot, may reveal the engagement and ease of the person during the task, whether or not a physical interaction is entailed \cite{Anzalone2015engagement,ivaldi2014frontiers,Chen2014NARStouch}.

The ability to estimate engagement and regulate social signals is particularly important when the robot interacts with people that have not been exposed to robotics, or do not have experience in using/operating them: a negative attitude towards robots, a difficulty in communicating or establishing mutual understanding may cause unease, disengagement and eventually hinder the interaction.

It seems therefore necessary to study how individual and social factors influence the issue of social signals during human-robot interaction, together with their relations to acceptance and engagement.

To evaluate the engagement during human-robot interaction, the most common metrics are based on the temporal dynamics of social signals, in particular gaze and speech \cite{Anzalone2015engagement,rich2010recognizing}.
The exchange of gaze (mutual and shared), the contingency of reactions to speech and gaze cues, the temporal dynamics of speech (utterance number, frequency, duration) are among the most common indicators of engagement during dyadic tasks \cite{ivaldi2014frontiers}.

However, there is evidence from the psychology literature that the dynamics of these social signals can be altered by individual factors \cite{LaFrance2004,Iizuka1992,Scherer1981}: we refer here to the set of behavioral, emotional, and cognitive tendencies that people display over time and across situations and that distinguish individuals from one another, such as personality traits and social attitudes.
The influence of personality traits on human behaviors during interactions with robots has been also documented in several studies \cite{Takayama2009proxemics,Dang2014personality,Aly2013personality}. 

Two individual factors seem particularly interesting for HRI: extroversion, a personality traits that is associated to positive emotions and social behavior \cite{BIGFIVE}, and negative attitude towards robots \cite{NARS2006}, a personal attitude that captures the projected anxiety of the person toward the interaction with a robotic device. 
Recent studies showed that there is a correlation between these traits/attitudes and the issue and dynamics of social signals, in particular gaze and speech \cite{Nomura2008}.
In this case, if they impact the issue of such social signals, they also affect the power of the metrics used as indicators of engagement.

Following this line of thought, the goal of this work is to study the relation between individual factors (extroversion and attitude toward robots) and the dynamics of gaze and speech produced by the human during an interaction with a robot (see Figure \ref{fig:concept}).

For this purpose, we designed a collaborative assembly task between a human and a robot. We made video and audio recordings (see Figure \ref{fig:setup}) of the interactions between the humanoid robot iCub and adult participants who previously submitted their questionnaires for evaluating the extroversion and negative attitude towards robots\footnote{In social psychology, there is a net distinction between personality traits and attitudes. Here, we use methods from differential psychology rather than social psychology: the distinction between the two is not important, as long as the two factors are two characteristics of the individual that are evaluated at a certain time prior to the interaction. We measured the attitude towards robots with the NARS questionnaire, a test that was created to capture the projected anxiety of the person before its interaction with the robot. We used it to evaluate an individual attitude prior to the direct interaction with the robot (participants filled the NARS questionnaire several days before the experiment - see details about the experimental procedure in Section \ref{section:protocol}).}. 
The questionnaire scores were later used to study the issue (frequency and duration) of utterances and gaze towards the robot issued by the human partner. Since our experiment also involved a physical contact between the robot and the person during the assembly, we distinguished between gaze towards the robot face and gaze directed towards the robot's hands, that perform the assembly thanks to the human guidance.

Our study shows that, at least for the cooperative assembly task, there is a correlation between extroversion score and the speech frequency and duration, while the negative attitude is related to the duration of gaze towards the robot. To summarize:
\begin{itemize}
\item \emph{the more one is extrovert, the more he/she will talk to the robot}
\item \emph{the more one has a negative attitude towards a robot, the less he/she will look at the robot face and the more he/she will look at the robot hands, where the physical interaction for the assembly takes place}
\end{itemize}

As gaze and speech are the main social signals used to evaluate engagement \cite{rich2010recognizing}, we provide significant results supporting the idea that engagement models used in HRI should take into account individual factors that can influence the production of such social signals.

By gaining a deeper understanding of the inter--individual factors that influence the exchange of gaze and speech during cooperative tasks, we aim at improving the design of robot controllers during social and physical interaction.
More generally, we would like to turn our findings into implications for the design of robot controllers that can adapt to the individual differences of the human partners.

\section{Background}\label{sec:background}

\subsection{Social signals: the building blocks for assessing engagement}

\begin{table*}
\centering
\begin{tabular}{|p{4.5cm}p{0.7cm}p{11cm}|}
\hline
Study & Ref & Social signals used to assess the engagement \\
\hline
\hline
Castellano et al., 2009 & \cite{Castellano2009} & Gazes towards the robot\\
 				& & Smiles\\
\hline
Ishii et al., 2011 & \cite{Iishi2011} &Gazes Towards the object the agent is talking about \\
			& &Gazes Towards the agent's head \\
			& &Gazes Towards anything else \\
\hline
Ivaldi et al., 2014 & \cite{ivaldi2014frontiers}  &Reaction time to the robot attention utterance stimulus \\
												& &Time between two consecutive interactions\\
\hline
Le Maitre and Chetouani, 2013 & \cite{Lemaitre2013} &Utterance directed to the robot \\
						& &Utterance  directed to self \\
\hline
Rich et al., 2010	& \cite{rich2010recognizing} &Gazes Focused (man and robot are looking at the same object \\
					& &Gazes Mutual (man and robot look at each other) \\
					& &Utterance Adjacent (two successive locutions, produced one by the robot, the other by the human, separated by a maximum interval) \\
					& &Utterance Responses (the subject responds to the robot through a gesture or a very short verbal intervention) \\
\hline
Sanghvi et al., 2011 & \cite{Sanghvi2011} & Postures (curve and inclination of the back) \\
\hline
Sidner et al., 2004 & \cite{sidner2004} &Gazes Shared (mutual or directed) \\
 			  & &Gazes Directed towards the robot without the latter looking at the human \\
\hline
Sidner et al., 2005 &  \cite{sidner2005} &Gazes Shared (mutual or directed) \\
 			  & &Gazes Directed towards the robot without the latter looking at the human \\
\hline
\end{tabular}
\caption{Social signals used in literature as metrics for the assessment of engagement.}
\label{table:literature}
\end{table*}

During interaction, a multitude of verbal and non-verbal signals are exchanged between the two partners. These so called \textit{social signals}  and their dynamics are the main bricks for the evaluation of the \textit{engagement} in HRI.

The engagement is defined as ``the process by which individuals involved in an interaction start, maintain and end their perceived connection to one another'' \cite{sidner2005}. As discussed in \cite{Anzalone2015engagement}, the engagement is related to the user experience, to the perceived control, feedback, interactivity, attention, and the fluctuations of the engagement during interaction are reflected into physiological changes and behavioral changes through verbal and non-verbal communication.

A social signal may be defined as ``a communicative or informative signal, or a clue which, directly or indirectly, provides an information about social facts, i.e. interactions, emotions, attitudes, valuations or social behaviors, social relations or identities'' \cite{Poggi2012}. The scope of social signals potentially extends to a large variety of behaviors and expressions: gestures, facial expressions, postures, gazes, etc. 
Anzalone et al. \cite{Anzalone2015engagement} partition the set of metrics for engagement evaluation into static and dynamic features. The first set comprises focus of attention and gaze analysis, head and body postural stability, with evaluation of pose and variance. The second set comprises joint attention, reaction times to attention cues, imitation, synchrony and rhythm of interaction. 

To assess the engagement during HRI experiments and tasks, researchers usually considers a subset of these social signals (see Table~\ref{table:literature}), frequently focusing on gaze and speech.

Gaze is one of the most important cues and carriers of information during the interaction.
It is indeed well established that mutual gaze and eye contact are crucial during human-human interaction \cite{Goffman1967}: the absence of eye contact at the right time, for instance at the end of a sentence, can be perceived as a withdrawal from the conversation and a sign of disengagement. 
Gaze in HRI can be analyzed differently depending on its direction and target.
For example, during verbal interaction \cite{rich2010recognizing,Iishi2011} or learning games \cite{ivaldi2014frontiers} it can be mutual (when the robot and the human partner look at each other) or directed/joint (when the robot and the human look at the same object or in the same direction).  
A third type of gaze can be the one directed by the human towards the robot, that the latter can return or not, depending on its joint attention skills \cite{sidner2004}.
 
Speech, and more specifically the dynamics of verbal exchange (e.g., turn-taking), is the other most important social signal for interaction, and it is a crucial indicator in the assessment of engagement \cite{Lemaitre2013,rich2010recognizing}. The metrics used for evaluating the engagement using this signal are for example the number, frequency and duration of utterances \cite{Iishi2011,rich2010recognizing}, the reaction time to utterance cues \cite{ivaldi2014frontiers}. Le Maitre and Chetouani \cite{Lemaitre2013} also proposed a qualitative distinction between language actions involving the locutions directed towards the robot, and those towards oneself.

Body language, which includes non-verbal behaviors such as facial expressions, posture and gestures, can also convey the intention and the engagement of the human partner. 
For example, Sanghvi et al. \cite{Sanghvi2011} analyzed the individual postures (the inclination and curve of the back) and their changes to assess the engagement of children playing chess against a humanoid robot.
The engagement was also studied in relation to positive facial expressions (e.g., smiles rather than grins) \cite{Castellano2009}, head movements such (e.g., nodding) \cite{Sidner2006} and gestures responding responding to a robot cue \cite{rich2010recognizing}.

To summarize, there are numerous studies that characterize the engagement and the interaction between humans and robots through the analysis of verbal and non-verbal signals. However \emph{gaze and speech are the most common social signals used to evaluate the engagement}, as clearly showed in Table \ref{table:literature}.

Since the engagement is a sort of emotional ``state'' of the human partner during the social interaction, and it may fluctuate over the interaction, it is interesting to study the  temporal dynamics of the social signals and the salient events associated to their evolution during the interaction. 
To estimate the engagement in HRI using the exchanged social signals, there are two main approaches in the literature.

The first approach consists in assessing the engagement of the human partner in real time. For instance, with a probabilistic approach Ishii et al. \cite{Iishi2011} demonstrate that a certain sequence of three gazing primitives (towards the object designated by the agent, towards the agent and towards any other direction) can reliably predict the human subject's withdrawal from an interaction. 
In their experiment, the robot was introducing a new model of mobile phone, and a sequence of gaze towards the robot then twice towards unrelated objects was linked to a disengagement.

In the second approach, that we may consider as ``global'', the engagement is neither measured in real time, nor on time intervals, but on the interaction as a whole. 
For instance, Sidner et al. \cite{sidner2004,sidner2005} suggest a metric combining the shared gazing time and the time spent by the subject looking at the robot for the whole duration of the interaction on one hand, and the assessment of the number of gazes that the participant returns to the robot during the same period of time on the other hand. 
With a similar approach, Rich et al. \cite{rich2010recognizing} developed a composite index defined by the average time intervals between two social connection events between the robot and the user in the course of an interaction, where the robot had to teach the participant how to prepare cookies. According to the authors, the events were divided into four sorts: 1) directed gaze, 2) mutual gaze, 3) adjacent utterances when two are produced in succession, one by the robot and one by the participant, with a maximum time gap between them, and 4) the replies to the robot with a gesture or a very short utterance. 
In the aforementioned studies, the researchers also observed the effect of various cues of the robot (e.g., robot nodding vs. not nodding) on the engagement of the user during the interaction.
A specific approach on the whole interaction was proposed by Le Maitre and Chetouani \cite{Lemaitre2013}: they proposed the ratio between the talking time directed towards the robot and the one towards oneself as indicator of engagement, with the rationale that an increased verbalization directed towards the robot can be interpreted as a stronger engagement (whereas the more the people talked to themselves, the lesser the engagement). 

To summarize, both considering the whole interaction and thin slices of interaction, measuring the engagement in HRI relies on the dynamics of the exchanged social signals, particularly gaze and speech.

However, there are no models that take into account context, task, social or individual factors that may affect the production of such signals, and subsequently the evaluation of the engagement. 

To the best of our knowledge, the HRI literature considering the inter-individual differences (concerning the personality) or the attitude (positive or negative) towards robots in the production of those signals is scarce. 
When discussing models of engagement, the human individual is considered as ``abstract'', expected to produce the same social signals at the same rhythm, despite any inter-individual difference that may affect the communication, the establishment and the continuation of the social interactions. 

It is however rational to consider that there can be personality traits, dispositions or attitudes that can make people talk, look and behave in a different way when facing the same interaction, especially with a device such as a robot.
For example, an introvert individual may talk less to or look less at the robot than an extrovert individual, without however being necessarily less engaged than the other. 
An individual with negative attitude towards robot may look less at the robot face, and look more at the robot's body, especially during close or physical interactions with the robot.
In short, the effect of  personality characteristics and of the attitudes towards robots could impact the dynamics of social signals, and subsequently undermine the metrics and models used in the literature to assess the engagement in HRI.

\subsection{Personality traits and attitudes}

As explained by Ajzen \cite{Ajzen1986}, ``attitudes and personality traits are latent, hypothetical dispositions that must be inferred from observable responses''. Their effect should be therefore observable on the overt actions of the individual.
The boundary between traits and attitudes is under debate; however it is acknowledged that both attitudes and personality traits influence our actions and behaviors, together with other social, contextual and individual factors \cite{Scherer1981}.
To make it simple, a personality trait is a characteristic of the human personality that leads to consistent patterns of behaviors, and is assumed to be almost invariant for an adult.
An attitude is a behavior tendency, directed towards people, objects, situations, and is generally determined by the social context, the background and experiences of the individual \cite{Wood2000}.

\subsubsection{Personality trait: extroversion}

The personality of an individual consists of several characteristics and dispositions, each  being described as a ``gathering of attitudes obviously linked to each other, or as patterns of cognitive treatment of the information or underlying psycho-physiological mechanisms generating specific dispositions towards some behaviors'' (\cite{Scherer1981}, p.116).

Among the existing personality models, the most well-known and studied is the Big Five \cite{NEOPIR1998}, which owes its name to the five traits descriptive of a personality: \emph{Extroversion}, \emph{Neuroticism}, \emph{Agreeableness}, \emph{Conscientiousness}, \emph{Openness to Experience}. This model is widely used in psychology to predict human behavior and cognition \cite{Wu2014,Rauthmann2012}, and is more and more also used in human-robot interaction \cite{Tapus08,Takayama2009proxemics}.

The extroversion dimension is the personality trait that notably (i) shows up more clearly during interaction, and (ii) has the greater impact on social behavior with respect to the other traits \cite{Zen2010}. It is linked to positive emotions, and identified through the tendency to be sociable, talkative, and self confident \cite{NEOPIR1998}. 
It seems to be fundamental to shape the way people interact \cite{Eysenck1981} and to establish and maintain social relations \cite{Wu2014}.
Beatty et al. \cite{Beatty2001} suggest that extroversion is one of the three major factors, together with neuroticism and psychoticism, that have some bearing on communication.  Moreover, it would also have an impact on the way individuals behave, and even on the quality of new social relations \cite{Berry2000}.

Although there is evidence in social psychology about potential links between the emission of various social signals (verbal and non-verbal) and the personality profile \cite{Argyle1976}, quantitative evidence is still needed. 
In particular, the current knowledge about extroversion and the issue of verbal and non-verbal signals is mostly limited to verbal dyadic and group interactions where there is typically no physical contact. 

Generalizing and characterizing the influence of individual differences and extroversion on verbal and non-verbal behaviors (e.g., gaze, head movements) is difficult \cite{LaFrance2004}; however, the literature in human-human interaction reports some evidence that the production of gaze and speech correlates to the level of extroversion of the individuals. 
For example, the level of extroversion has an effect on the frequency and duration of gazes towards a person during face-to-face conversations \cite{Iizuka1992}: extroverts gaze longer than introverts.
In a similar way, Wu et al. \cite{Wu2014} showed that extrovert individuals tend to focus their attention on the area of the eyes on pictures of human beings longer than introverts. 
The influence of personality traits, especially extroversion, on the gaze is also reported for non-social tasks such as fixating abstract images \cite{Rauthmann2012}. 

With regards to verbal communication, Costa et al. \cite{NEOPIR1998} noted that one of the most clear signs of extroversion for an individual is to be more talkative, which also leads to a lesser number of pauses during conversation \cite{Scherer1981}. Extrovert people would also tend to use shorter sentences at an increased rate than introvert people in informal situations involving another language \cite{Dewaele00}. 
The link between extroversion and speech dynamics was exploited for automatic classification of personality from videos of interaction between small groups of people: in \cite{Pianesi08,Lepri2010} the authors showed that the talking and interaction timing ratio are positively correlated to the level of extroversion.

To summarize, there is evidence from the literature on the influence of the extroversion trait on the dynamics of gaze and speech in human-human interaction. This certainly biases the current metrics and models for assessing engagement, that do not take into account such individual factors \cite{Anzalone2015engagement,rich2010recognizing}. 

Extending such studies to human-robot interaction, with the variability of tasks, situations and robots, it is certainly challenging. In this paper, we provide evidence that the dynamics of gaze and speech is related to the extroversion during a human-robot assembly task.

\subsubsection{Negative attitude towards robots}

As the literature seems to allege, extroversion may bring up inter-individual communication differences during social interactions between humans. While aversion towards other people may be identified through the personality models, there is currently no model that allows us to assess the dislike of technology, and more specifically robots. An individual may appear to be very sociable, while very wary of technology. 
For robots, this evaluation seems particularly critical. Currently, robots are diffused in factories and service and are mostly used or operate by skilled people that received some robotics training (i.e., experts). However, robots are gradually becoming available and accessible outside the classical settings, to ordinary people that have not received any robotics training (i.e., non-experts). Ordinary people without a proper knowledge of the platform are not typically aware of the limits and the real capabilities of the robots, because of their lack of prior experience with them and frequently limited background knowledge. 
Some people might be technophobic, some might have developed an anxiety towards robots, influenced by some recent trends in the public media\footnote{See for example the press article: ``Will workplace robots cost more jobs than they create?'' \url{http://www.bbc.com/news/technology-27995372}}, some may be influenced positively or negatively by movies\footnote{We interviewed our participants after the experiments. Some reported that they ``do not like robots because they are going to take our jobs''. Some reported to have enjoyed the experiment with the robot and made explicit reference to their expectations being influenced by ``the robots of Star Wars''.} and literature \cite{Mara2015}. This \textit{a priori} may reflect in differences in their behavior and communication, and not be dependent necessarily by their personality traits.

It seems therefore necessary to take into account a personality characteristic that is related more to technology rather than human beings, and more particularly to social robots and humanoids.

This category of robots has been recently studied to better understand the reasons that may cause negative attitudes towards this ``too human-like'' technology \cite{Saygin2012}. 
The most known negative effect linked to the robot appearance is the so called ``Uncanny Valley'' effect: described by Mori in 1970, it describes the fact that a robot excessively ``human-like'' arouses a sense of unease and repulsion, whereas robots with a moderate level of human likeness or humanoids that can be clearly identified as machines arouse more affinity \cite{UncannyRAM2012}.
While numerous studies show that the humanoid appearance is accountable for opinions and attitudes towards the robots \cite{Gray2012}, other factors also seem to affect these attitudes: movements speed and profiles, distance during the interaction, voice and temporal dynamics of verbal exchanges between the human and the robot. 
From a methodological point of view, attitudes towards the robots are usually assessed through  free verbalization (e.g., interviews) and attitude scales.
Nomura and colleagues \cite{Nomura2006nars,NARS2006} developed a questionnaire for the valuation of negative attitudes towards humanoid robots: the Negative Attitude towards Robots Scale (NARS). In a series of studies, they could demonstrate the effect of a negative attitude towards robots on the communication, in particular on the time of the verbal response, which increases with the more the negative attitude of an individual.

It appears that a negative attitude towards robots has therefore an impact on the way people interact verbally with a robot. Someone with a more negative attitude towards robots may talk less to the robots: this could be misinterpreted as a sign of disengagement.
Since speech dynamics is one the main indicators for engagement assessment, it should be recommended to take into account the impact of attitudes in the models for assessing the engagement based on the interpretation of social signals emitted by the human during HRI.

Incidentally, the influence of the negative attitude towards robots on social signals has been studied during interaction tasks with a significant verbal component, but not yet in tasks with physical interaction. 
However, since this attitude captures the worry of the person projected towards an interaction with a robot, we expect that its influence on the social signals will be more visible in tasks with contacts between the robot and human. 
In this case, the close proximity with the robot and the touch should highlight the unease and anxiety of the human.
This effect was observed by Chen et al. in the robot nursing experiments \cite{Chen2014NARStouch}, where the authors showed that people with negative attitude towards robots responds less favorably to robot-initiated touch. 
Our intuition is that touching the robot in particular should produce more distress, therefore making the humans gaze more at the body parts where the interaction occurs.

\begin{figure*}[ht]
\centering
{
\includegraphics[width=0.7\hsize]{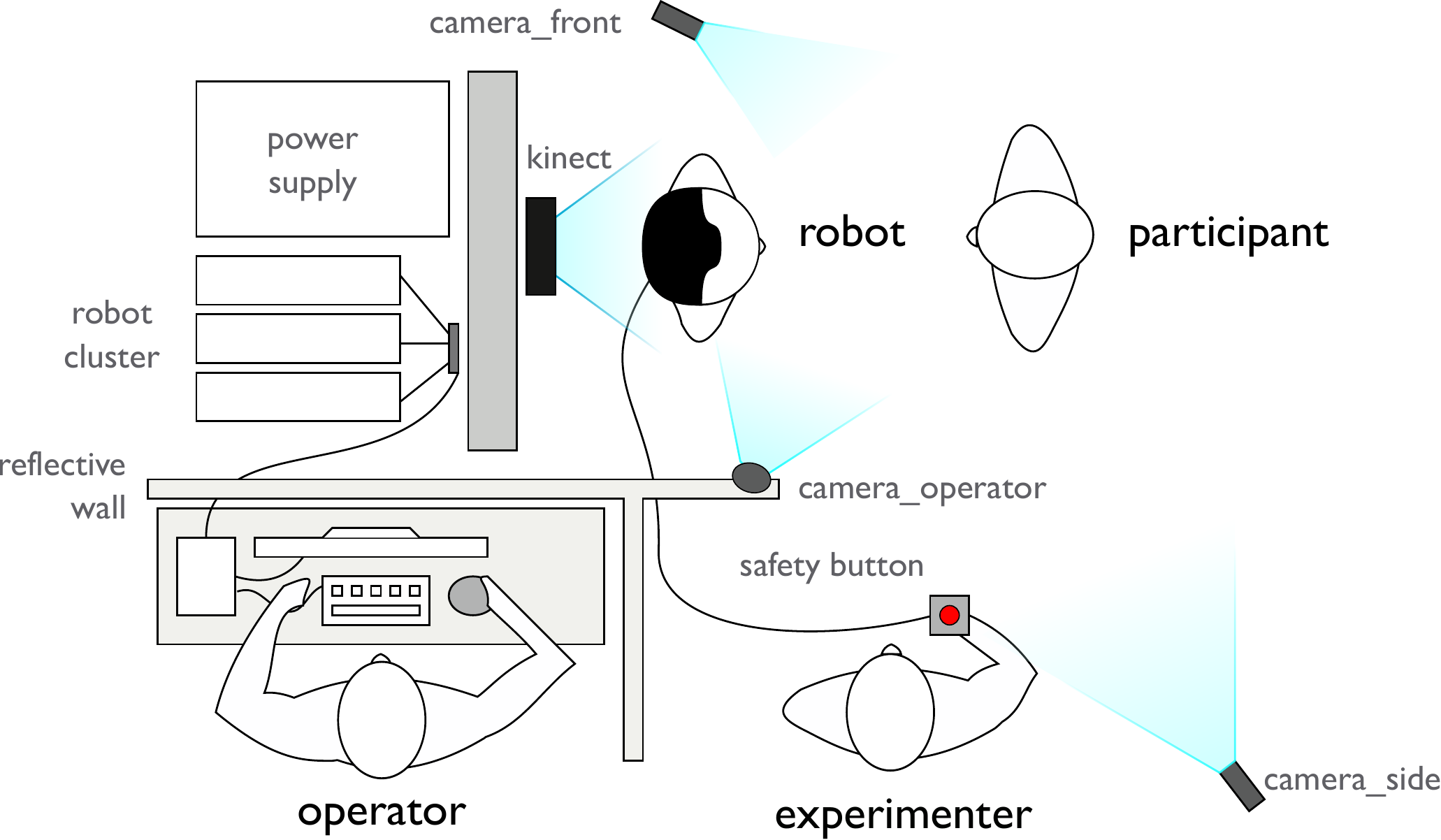}
}
\caption{The experimental setup. The participant is standing in front of the robot iCub; their interaction is recorded by a Kinect, two standard HD cameras (front and side view of the scene). The experimenter monitors the interaction from the side, not too far but close enough to be able to push the safety button and intervene in case of emergencies. The operator is hidden behind a wall, and he controls the robot monitoring the interaction through a webcam placed over the robot. The power supply and cluster of the robot are hidden behind a cabinet.}
\label{fig:setup}
\end{figure*}

\section{The study}

\subsection{Rationale}

There are several studies on the influence of individual factors on the production of social signals during human-human interactions (for example, \cite{Lepri2010,Vinciarelli14}). 
Recent studies on the link between personality traits and social signals have also appeared in the HRI community (for example, \cite{Tapus08,Aly2013personality}). 

However, to the best of our knowledge there is no study yet examining the relation of individual factors to gaze and speech during an assembly task. 
In this type of cooperative tasks, the interaction between the human and a robot entails a physical and a social dimension. The contact with the robot (at the level of the hands, in this case) and the close proximity between the partners may induce variations of the production of gaze and speech with respect to simple face-to-face interactions with a predominance of verbal exchange. The alterations of the dynamics of the signals could be due to the task and/or to some characteristics of the individual, for example its personality or attitude towards robots.

The engagement models do not currently differentiate between tasks with or without contact, and do not take into account individual factors that may induce changes in the dynamics of social signals.

It is therefore necessary to provide evidence of the relation between these elements to improve the classical models of engagement. We do it in this paper for a dyadic task that is fundamental for robotics in service and industry: the cooperative assembly.
Furthermore, it seems necessary to take a comprehensive approach with respect to the individual factors, considering personality traits and attitudes towards robots, as the personality traits alone could not be sufficient to explain the variation of the social signals during an interaction with a robot.

\subsection{Research hypotheses}\label{sec:hypotheses}

Based on the literature review discussed in Section \ref{sec:background}, we expect that participants that have high scores of extroversion will talk more to the robot; we also expect that participants with a very high negative attitude towards robots score will avoid gazing at the robot. 
Due to the specificity of the task, involving a contact between the human and the robot, we expect that participants with a high negative attitude towards robots will gaze more at the robot hands (area of contact between the human and the robot).

Therefore, we pose five research hypotheses:

\medskip
\textbf{Hypothesis 1}: 
\textit{If the extroversion dimension is related to the frequency and duration of utterances addressed by the human to the robot, then we should find a positive correlation between the questionnaire score of extroversion and these variables.}

\textbf{Hypothesis 2}: 
\textit{If the extroversion dimension is related to  the frequency and duration of gazes directed towards the robot's face, then we should find a positive correlation between the questionnaire score of extroversion and these variables. }

\textbf{Hypothesis 3}: 
\textit{If the negative attitude towards robots is related to the frequency and duration of the utterances addressed by the human to the robot, then we should find a negative correlation between the questionnaire score of the negative attitude towards robots and these variables.}

\textbf{Hypothesis 4}: 
\textit{If the negative attitude towards robots is related to the frequency and duration of gazes directed towards the robot's face, then we should find a negative correlation between the questionnaire score of the negative attitude towards robots and these variables.}

\textbf{Hypothesis 5}: 
\textit{If the negative attitude towards robots is related to the frequency and duration of gazes directed towards the areas of contacts between the human and the robot, then we should find a positive correlation between the questionnaire score of the negative attitude towards robots and these variables.}

\medskip
The hypotheses were tested through an interaction task where human participants had to cooperate with the humanoid robot iCub \cite{icub2013} to assemble an object. We made video and audio recordings of the interactions between the humanoid iCub and adult participants who previously submitted their questionnaires for evaluating the extroversion and negative attitude towards robots.\footnote{In social psychology, there is a net distinction personality traits and attitudes. Here, we use methods from differential psychology rather than social psychology. We measured the attitude towards robots with the NARS questionnaire, a test that was created to capture the projected anxiety of the person \textit{before} its interaction with the robot. We used it to evaluate an individual attitude prior to the direct interaction with the robot (participants filled the NARS questionnaire several days before the experiment - see details about the experimental procedure in Section \ref{section:protocol}).}

This task was part of a set of experiments within the project ``Engagement during human-humanoid interactions'' (EDHHI)\footnote{\url{http://www.loria.fr/~sivaldi/edhhi.htm}}, to investigate the acceptance, engagement and spontaneous behavior of ordinary people interacting with a robot.
The experimental protocol used in this work (Ivaldi et al., ``Engagement during human-humanoid interaction'', IRB n.20135200001072) received approbation by the local Ethics Committee (CERES) in Paris, France.

\section{Materials and methods}\label{sec:material}

\subsection{Questionnaires}

To evaluate the extroversion and the attitude towards robots of the participants, we used two questionnaires: the Revised Personality Inventory (NEO-PIR) \cite{NEOPIR1992} and the Negative Attitude towards Robots Scale (NARS) \cite{NARS2006}.

The first is used to assess the personality traits according to the Big Five model \cite{BIGFIVE}. The official French adaptation of the questionnaire was used \cite{NEOPIR1998}. We retained only the questions related to the assessment of the extroversion dimension, that is 48 questions divided into six facets: Warmth, Gregariousness, Assertiveness, Activity, Excitement seeking and Positive emotions\footnote{We cannot report the questions, as the questionnaire is not publicly available: we refer the interested reader to the English manual \cite{NEOPIR1992} and the official French adaptation that we used \cite{NEOPIR1998}.}.
The order of the questions followed the original questionnaire; answers were on a Likert-type scale from 1 (Totally disagree) to 5 (Totally agree).

The second questionnaire consists of 14 questions divided in three sub-scales: ``Negative attitude towards situation of interaction with robots'' (S1), ``Negative attitude towards social influence of robots'' (S2) and ``Negative attitude towards emotions in interaction with robots'' (S3). The order of the questions followed the original questionnaire; answers were on a Likert-type scale, from 1 (Strongly disagree) to 7 (Strongly agree).
To the best of our knowledge, an official French adaptation of the NARS questionnaire does not yet exist. 
For the experiments, we therefore proposed our French adaptation of the NARS questionnaire, taken from \cite{Nomura2006nars}. 
Our questionnaire was produced with a double translation made by three different researchers, fluent in both English and French, and was validated by a group of ten external people to ensure that the French translation was properly understood\footnote{A recent paper from Dinet \& Vivian \cite{NARSfrench} studied the NARS and validated it on a sample of French population. Their study was published only after our work and experiments. They employed their own translation of the questionnaire, which has some slight differences with ours, mostly due to some \textit{nuances} of the French language. These do not preserve the original meaning when translated back into English. In their paper there is no mention of a double translation mechanism for validating the French adaptation of the questionnaire.}.
We report the questions in both French and English in Table \ref{table:nars} in Appendix \ref{appendix:nars}.

The participants also filled up a post-experimental questionnaire for subjective evaluation of the assembly task with the robot.
The questionnaire was designed to catch the impressions and feedback of the participants about the task, their interaction experience and in particular the way they perceived the physical interaction with the robot. 
We report the questions in both English and French in Table \ref{table:postexperimentquestionnaire} in Appendix \ref{appendix:postexpquestionnaire}. 
The order of the questions followed the original questionnaire; answers were on a Likert-type scale from 1 (Totally disagree) to 7 (Totally agree).

\subsection{Experimental setup}\label{sec:experimentalsetting}

The experiments were conducted in the Institut des Systèmes Intelligents et de Robotique (Paris, France), in the laboratory room of the iCub robot. 

The experimental setup was organized as depicted in Figure \ref{fig:setup}. 
The robot was standing on a fixed pole, so that it could not fall.
The robot was semi-autonomous, i.e., it was controlled by an operator hidden behind 
a reflective wall (a plastic divider with reflective surface), built  
to prevent the participants to see the operator and the experimenter, while giving the experimenter the possibility to monitor the interaction and intervene promptly in case of problems\footnote{This was done as a safety measure. However, nothing happened during the experiments: the experimenter never had to push the safety button, and she never had to stop the physical interaction between the robot and the subject.}. 

Two cameras were recording the participants, as shown in Figure \ref{fig:setup}. 
One camera was placed behind the robot on its left side, in such a way to observe the human face and upper-body during the close interaction with the robot, while the other one was placed laterally to take the scene as a whole.

\begin{figure}[ht]
\centering
{
\includegraphics[width=0.5\hsize]{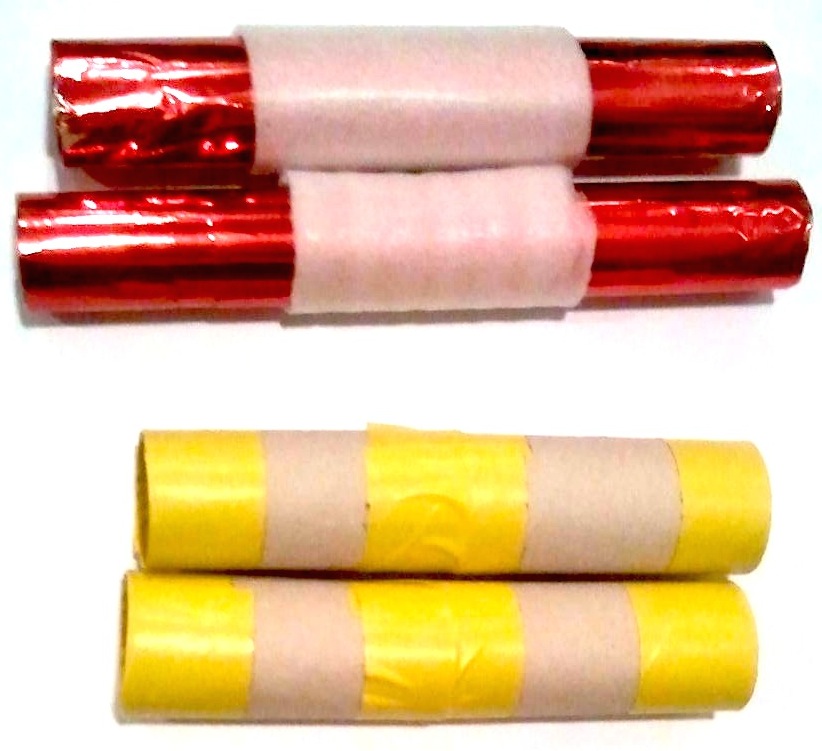}
}
\caption{Colored paper rolls used in the assembly task.}
\label{fig:rolls}
\end{figure}

The colored rolls used for the assembly task are shown in Figure \ref{fig:rolls}.

The experiments were carried out with the humanoid robot iCub \cite{icub2013}. The robot is approximately 104 cm high, weights about 24 kg, and has the shape of a 4 years old child.

To facilitate the control of the robot by the operator, we developed a graphical user interface (GUI) to quickly send high-level commands to the robot in a wizard-of-Oz mode (WoZ). 
The operator was constantly monitoring the status of the robot, and could intervene to send high-level or low-level commands to the robot, in prompt response to unexpected actions or requests of the participants, using a dedicated graphical interface (see Appendix \ref{appendix:GUI}).

The robot was always controlled in impedance \cite{idyn2012}, to make it compliant in case people would touch it accidentally or intentionally before the construction task. 
When people had to physically manipulate the robot to move its arms and accomplish the task, the operator was switching the robot into a zero-torque control mode that allowed the arms to be driven lightly by the participants. 
For safety issues, the operator could stop the robot motion at any time simply switching the robot to position control, and at the same time the experimenter monitored the whole interaction 
and was able to intervene and stop the robot in case of urgency at any time using the robot safety button. 
Facial expressions and speech were enabled (more details in Appendix \ref{appendix:GUI}).
The robot always assumed the same neutral/positive expressions, to avoid confusing the participant or suggest an underlying 
robot ``emotional status''.

\subsection{Participants} 

Prospective participants were recruited through a generic announcement for HRI studies, diffused on a mailing-list. 
Participants that volunteered in the study received a 10 euros voucher as a symbolic reimbursement for travel expenses. They signed an informed consent form to partake in the study and granted us the use of their recorded data and videos.
N=56 voluntary healthy adults took part in this study: 37 women, 19 men, aged 19 to 65 (mean=36.95, $\sigma$=14.32). The participants were all native French speakers.

\subsection{Experimental procedure}\label{section:protocol}

After volunteering to take part in the experiment, the participants received an ID number to preserve anonymity during the study. 
The personality traits of the participants were retrieved by questionnaires that were filled up through an online web form two weeks before doing the experiment, to avoid influences of the questions on their behavior.

The day of the experiment, participants were welcomed by the researcher and informed about the overall procedure before signing an informed consent form granting us the use of all the recorded data for research purposes.

Before the experiment, the participants had to watch a short video presenting the iCub, its body parts and some of its basic movements\footnote{It is a dissemination video from IIT showing the iCub, available on Youtube: {http://youtu.be/ZcTwO2dpX8A}.}. The video did not provide any information about the experiments. It was instrumental to make sure that the participants had a uniform prior knowledge of the robot appearance (some participants may have seen the robot before on the media).

After the video, each participant was equipped with a Lavalier microphone to ensure a clear speech data collection, then was introduced to the robot.
The experimenter did not present the experimental setup (e.g., show the location of the cameras) except showing the robot, and she did not provide any specific instruction to the participants about what to do or say and how to behave with the robot. 
Most importantly, she did not say anything about the 
fact that the robot was not fully autonomous:
since the operator was hidden behind a wall, mixed with other students of the lab, the participant had no cue that the robot was controlled by someone else\footnote{In the post-experiment interview, we asked the participants if they thought or had the impression that the robot was controlled by someone: all the participants thought that the robot was fully autonomous.}.
The robot was in a standing position, gently waving the hands and looking upright, while holding a colored toy in its right hand. 
Once the participants were standing and looking in front of the robot, they were free to do whatever they wanted: talk to the robot, touch it, and so on. 

\begin{figure*}
\centering
\includegraphics[width=0.95\hsize]{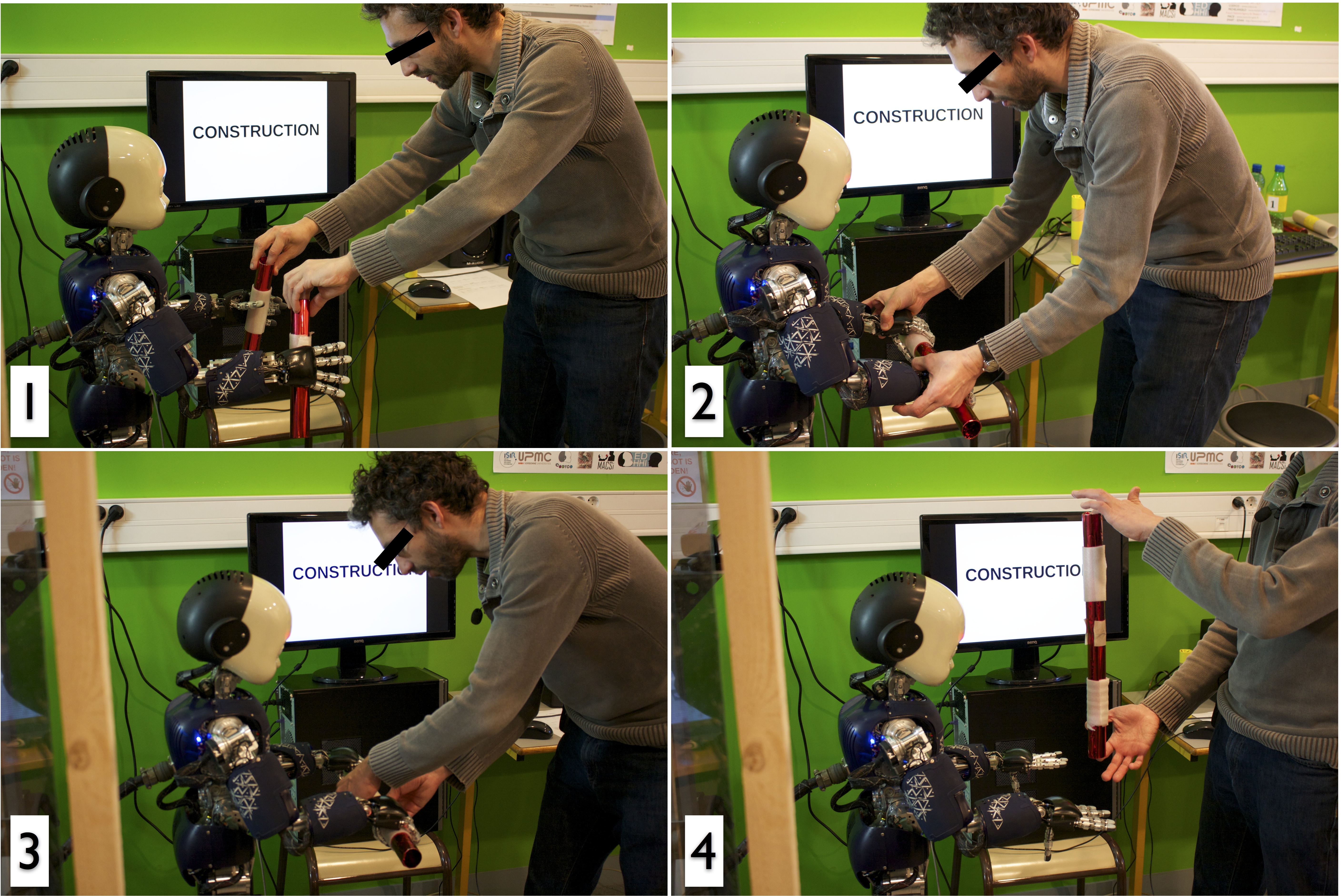}
\caption{Demonstration of the assembly task: 1) the participant asks the robot to grasp the two cylinders; 2) the participant grabs the robot arms and demonstrates how to move them to align the two cylinders; 3) the participant fixes the cylinders with some tape while the robot is holding them; 4) the participant retrieves the assembled object from the robot. }
\label{fig:task}
\end{figure*}

The experimenter explained that the goal of the task was to create an object in collaboration with the robot.
To create the object, they simply had to assemble two paper rolls and fix them with some tape.
The participant could grab the robot arms to demonstrate the bi-manual movement necessary to align the two rolls, as shown in Figure \ref{fig:task}. 

\begin{figure}
\centering
\includegraphics[width=0.95\hsize]{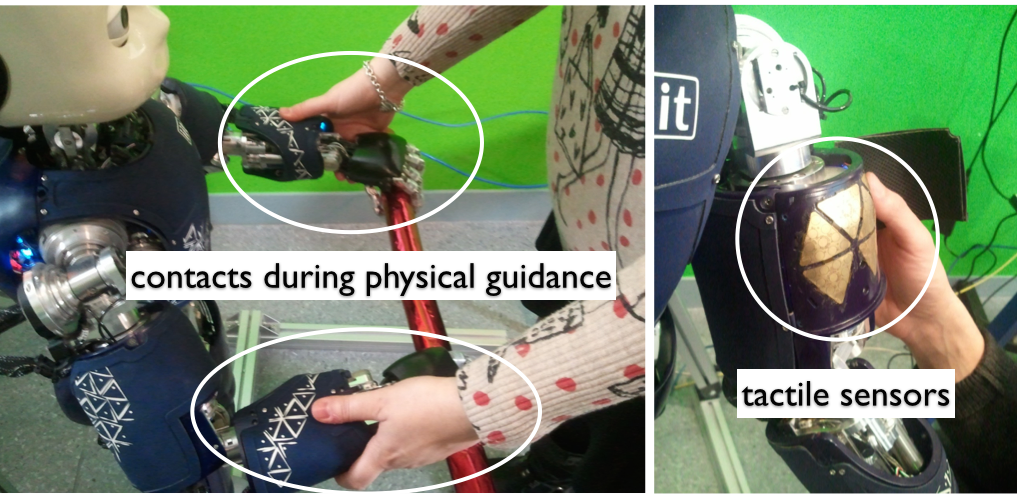}
\caption{Demonstration on how to safely grab the robot arms for kinesthetic teaching in the assembly task: the hands of the experimenter grasp the robot forearms on a part covered by the skin. On the left, the distributed tactile sensors underneath the cover.}
\label{fig:contacts}
\end{figure}

As the task required a physical interaction with the robot, for safety reasons the experimenter had to provide a short demonstration to show the participant how to grab the robot arms in a safe manner and how to ``move'' the robot arms by guidance to teach the robot a desired movement\footnote{The demonstration was also part of the safety measures required by the Ethics Committee to approve our protocol.}.
This demonstration was necessary to make sure that the participants would grab the robot forearm on the cover parts covered by the skin, for their own security and to prevent damaging of cables and robot hands (see Figure \ref{fig:contacts}).
All the participants received the identical demonstration. 
To show a movement to the robot, the experimenter gently grasped the robot forearms touching the skin and saying ``\textit{Be compliant}''. The robot operator then switched the control mode of the robot arms to zero-torque control, so that the experimenter could gently move the arms. To make the arms hold the position, the experimenter said ``\textit{Hold on}''. The operator then switched the control mode of the arms to impedance position control\footnote{The operator could switch the control mode without the need of the verbal command, since he had a direct visibility of the interaction zone in front of the robot through an additional camera that was centered on the workspace in front of the robot (see Figure \ref{fig:setup}).}. 

The short demonstration was necessary for safety reasons, because the participants were not robotics experts.
The experimenter precised that the demonstration was not to be used as a template on how to perform the task with the robot, as neither the task nor the interaction were scripted and the robot would follow the participant's guidance and commands.

To accomplish the assembly task, the experimenter precised that it was necessary to explain  to the robot how to realize the assembly step by step, even if no scripted procedure was provided. No explicit instructions were given to the participants on how to explain the procedure to the robot.

We remark that the interaction between participant and robot was not scripted, and our aim was to let it be as much as spontaneous as possible for a first human-humanoid interaction.

The experimenter then gave the participants the first two colored paper rolls and invited the participant to start the assembly task with the robot; the task had to be repeated three times with three pairs of paper rolls, so as to build three objects. The paper rolls and the tape were conveniently placed on a table next to the participants.
The participant was free to start at his/her own convenience, and to make each trial last how much he/she wanted to.
Some paper rolls used in the experiments are shown in Figure \ref{fig:rolls}.

Once the participants finished the assembly task, repeated three times, the experimenter led the participant back to a computer to make him/her fill a post-experiment questionnaire and then get feedback and impressions through a short interview.

\begin{figure*}
\centering
{
\includegraphics[width=0.24\hsize]{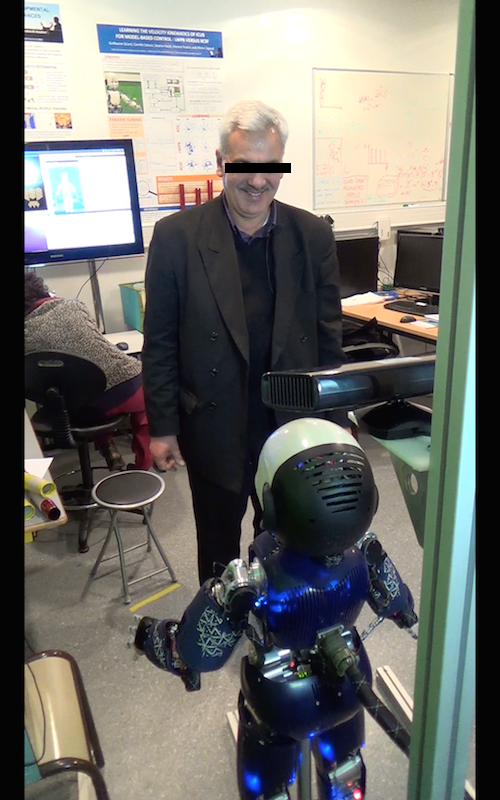}
\includegraphics[width=0.24\hsize]{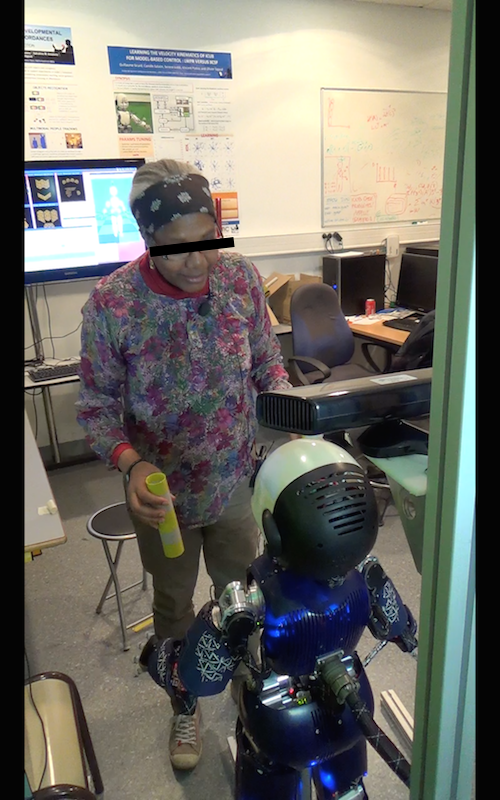}
\includegraphics[width=0.24\hsize]{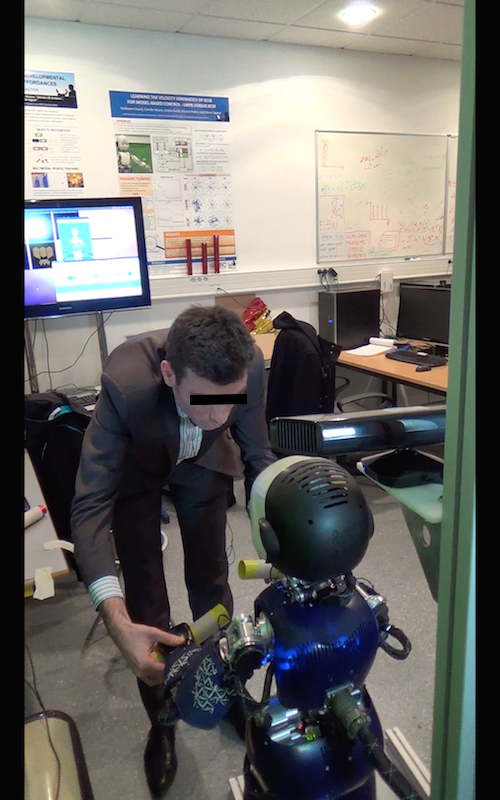}
\includegraphics[width=0.24\hsize]{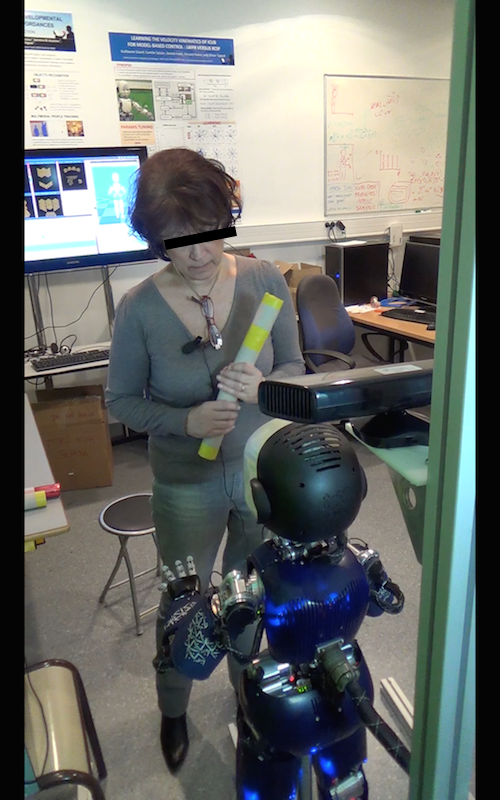}
}

\caption{Some participants gazing at the robot face. From left to right: when the participant meets the robot, handing the cylinders, during the alignment of cylinders, and when the object is built.}
\label{fig:shotsgaze}
\end{figure*}

\begin{figure*}
\centering
{
\includegraphics[width=0.25\hsize]{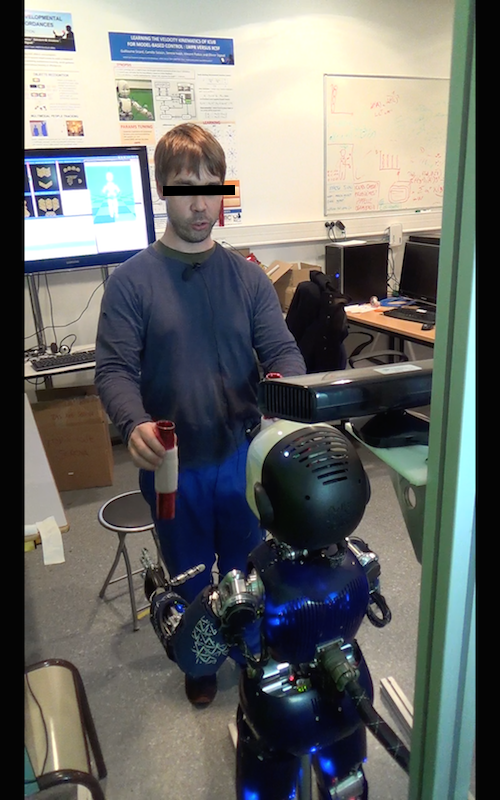}
\includegraphics[width=0.25\hsize]{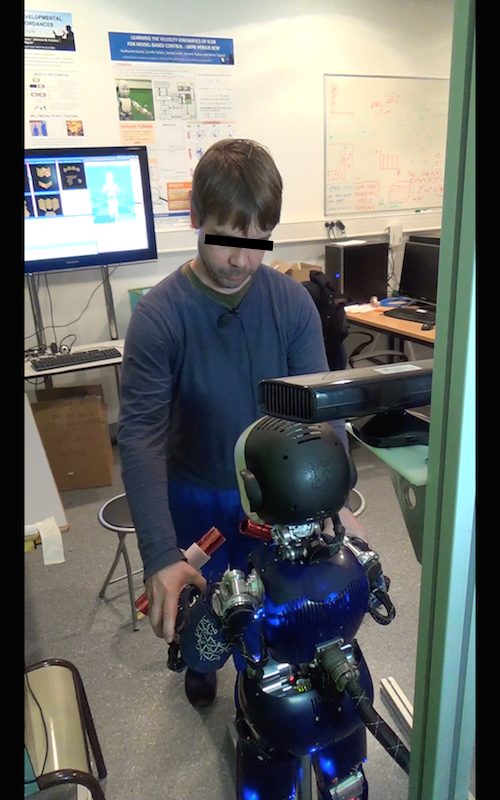}
\includegraphics[width=0.25\hsize]{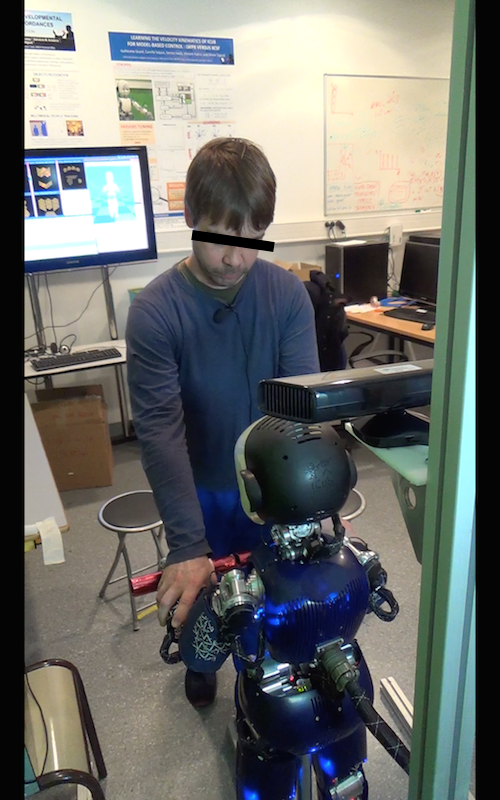}

\includegraphics[width=0.25\hsize]{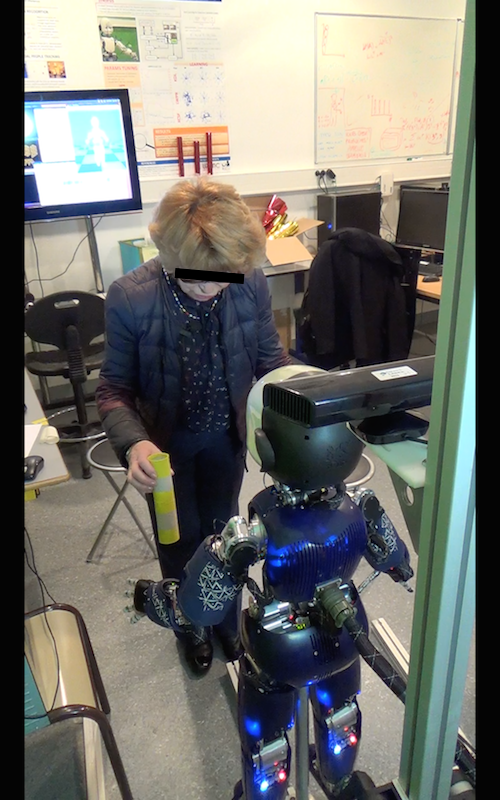}
\includegraphics[width=0.25\hsize]{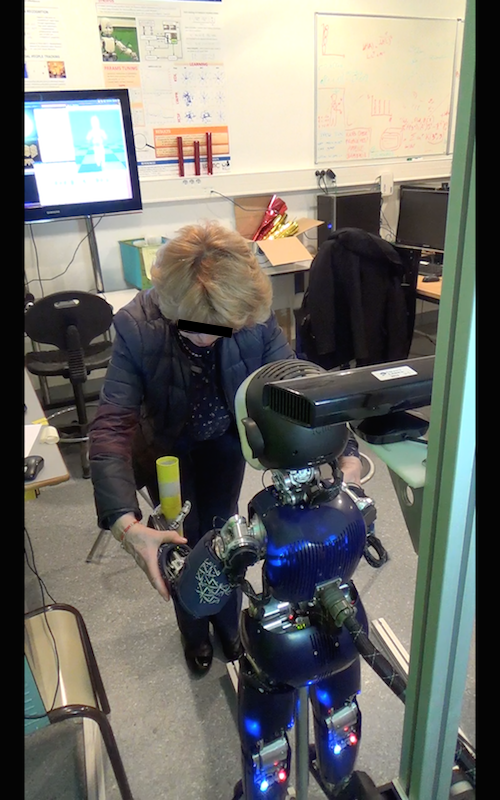}
\includegraphics[width=0.25\hsize]{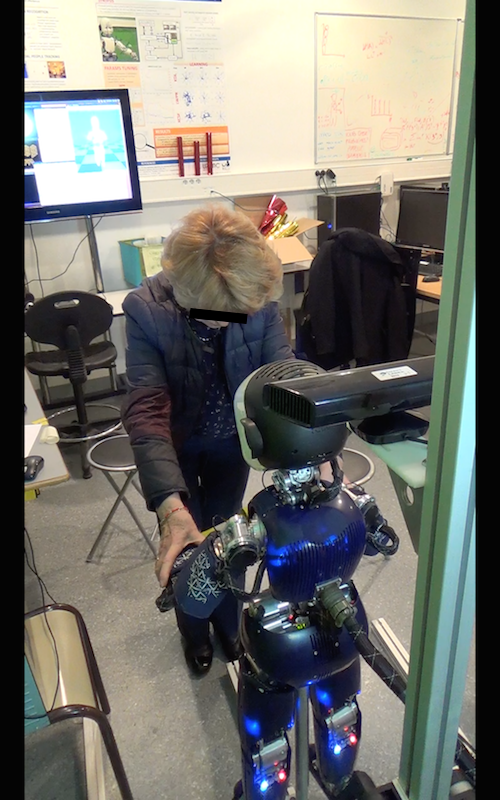}
}

\caption{Some participants performing the assembly task (screenshots from the front camera). The three images show the participants giving the cylinders to the robot (left), grabbing the robot arms (center) then moving the arms to align the cylinders (right).}
\label{fig:shotscontact}
\end{figure*}

\begin{figure*}
\centering
{
\includegraphics[width=0.24\hsize]{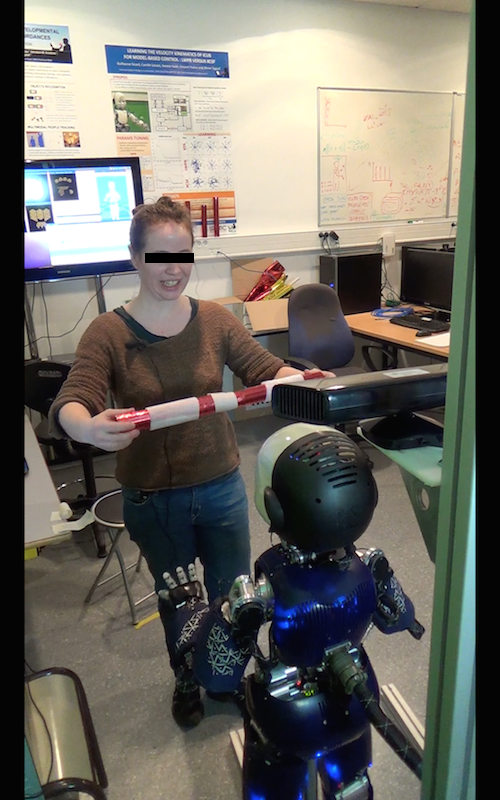}
\includegraphics[width=0.24\hsize]{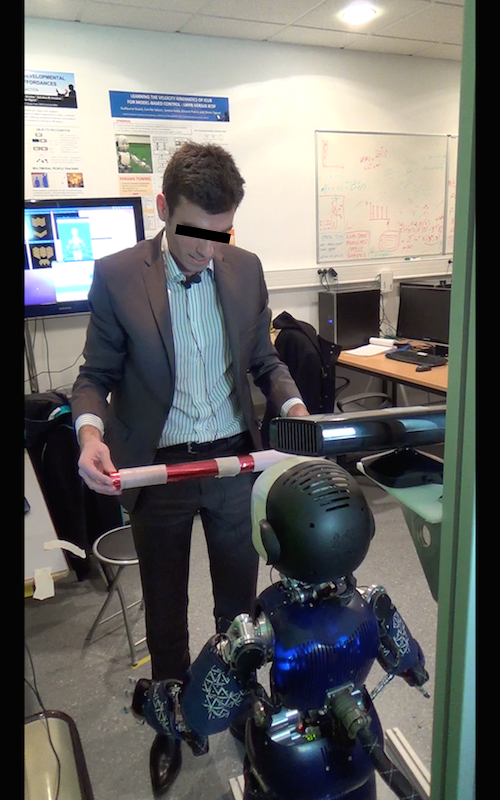}
\includegraphics[width=0.24\hsize]{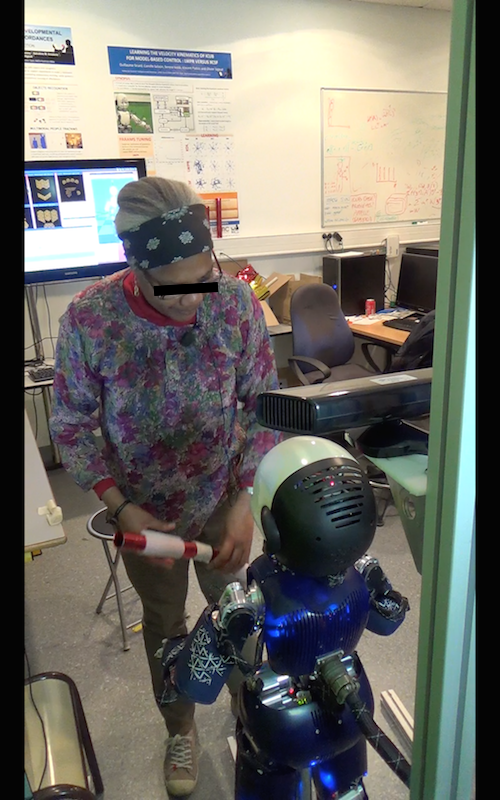}
\includegraphics[width=0.24\hsize]{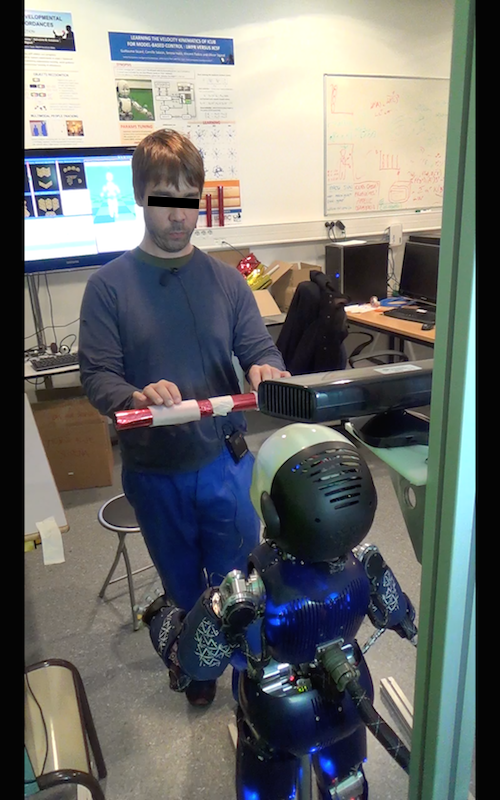}
}
\caption{Some participants showing the final object to the robot, after the collaborative assembly.}
\label{fig:shotsfinished}
\end{figure*}

\begin{table*}
\textbf{Post-experimental questionnaire for human-humanoid collaborative tasks with physical interaction}

\begin{tabular}{|p{12cm}|p{3.5cm}|}
\hline
 Questionnaire Item & Subjective evaluation (score mean $\pm$ stdev) \\
 \hline
 \hline
\multicolumn{2}{|l|}{\textbf{Questions related to the task}} \\
\hline            
 The assembly task was easy to do. & 5.49 $\pm$ 1.39\\
 \textbf{The assembly task was interesting to do.}& \textbf{5.75 $\pm$ 1.61}\\
 Someday I could work with this robot to build something of interest.& 5.03 $\pm$ 1.67\\
\textbf{Someday I could work with a robot to build something of interest.}& \textbf{5.87 $\pm$ 1.07}\\
\hline
\multicolumn{2}{|l|}{\textbf{Questions related to the physical interaction (e.g., touching the robot)}} \\
\hline  
\textbf{I was worried to must touch the robot to assembly the objects with it.} & \textbf{2.13 $\pm$ 1.46}\\
\textbf{I was afraid to touch the hands of the robot.} & \textbf{2.36 $\pm$ 1.72} \\
 I was afraid to damage the robot. &	3.57 $\pm$ 1.91 \\
\textbf{The robot does not look dangerous.}& \textbf{6.00 $\pm$ 1.57}\\
\textbf{The robot is not threatening.}& \textbf{6.02 $\pm$ 1.49}\\
\hline
\multicolumn{2}{|l|}{\textbf{Questions related to the cognitive/social interaction}} \\
\hline 
 During the assembly, I would have preferred that the robot tells me what it thinks, if it understands well. & 5.19 $\pm$ 1.61 \\
 The robot understood what I explained to it.&	 5.38 $\pm$ 1.39\\
 The robot should be more reactive.& 4.65 $\pm$ 1.56\\
  The robot was nice. & 5.49 $\pm$ 1.37\\
\hline
\multicolumn{2}{|l|}{\textbf{Questions related to the robot features}} \\
\hline 
 The robot moves its head too slowly. & 3.32 $\pm$ 1.41\\
 The robot moves its arms too slowly. & 3.55 $\pm$ 1.33\\
\textbf{The facial expressions of the robot trouble me.}& \textbf{2.03 $\pm$ 1.29}\\
 The voice of the robot is pleasant.& 4.51 $\pm$ 1.84\\
\hline
\end{tabular}
\caption{The scores of the post-experimental questionnaire for evaluating the perception and interaction with the iCub in the assembly task of this work. The second column reports the mean and standard deviation of the scores attributed on a 7-items Likert scale (from 1=totally disagree to 7=totally agree) by the N=56 participants in this study. We highlight in bold the questions where the score is close to the maximum or the minimum score.}
\label{table:postexperimentquestionnairescores}
\end{table*}

\subsection{Data analysis}

The questionnaires scores for extroversion and NARS were computed according to their authors' recommendation. 

The audio-video recordings were analyzed with CowLog software \cite{CowLog2009}. 
 Six events were annotated: beginning of the interaction, end of the interaction, beginning of a gaze by the participant towards the robot's face or hands (i.e., the contact area), end of that gaze, beginning of an utterance addressed to the robot, end of that utterance. 
The gaze direction was approximated by the head orientation, as it is often done in literature \cite{ivaldi2014frontiers,Ba2009}. 
We considered two consecutive utterances whenever there was a pause of at least 500ms. 

We computed from the events' timestamps the following six dependent measures: frequency and duration of gaze towards the robot's face, frequency and duration of gaze towards the robot's hands, frequency and duration of utterances addressed to the robot. 
These indicators were normalized by the total duration of the interaction, to take into account inter-individual variability in terms of task duration.

We used Pearson's correlation to test of correlation of the extroversion and attitude towards robots on the frequency and duration of gaze and utterances\footnote{Correlation is frequently used to study the link between personality and behavior, as discussed in \cite{LaFrance2004}, a survey on the link between extroversion and behavior where all the cited studies use correlations to test their hypothesis.}.

\section{Results}

The average time to complete the task was 246.10s ($\sigma$= 75.45). 
On average, the participants talked to the robot for 69.92s ($\sigma$=38.38), addressing to it 57.54 utterances ($\sigma$= 25.65); they looked at the robot's face for 42.55s ($\sigma$=29.25), gazing at the face of the robot 12.13 ($\sigma$=6.57) times; they looked at the robot's hands for 162.46s ($\sigma$=57.14), gazing at the hands 11.30 ($\sigma$=5.70) times.

\subsection{On the individual factors}
To ensure that the two questionnaires capture two different individual factors, we computed the correlation between the scores of extroversion and negative attitude towards robot obtained by our population of participants.
We did not find a significant correlation between the two ($r^2$=-0.213; p=N.S.), neither between extroversion and  
each of the three sub-scales: negative attitude towards interaction with robots ($r^2$=-0.156; p=N.S.), negative attitude towards social influence of robots ($r^2$= -0.156; p=N.S.), and negative attitude towards emotions during the course of interactions with robots ($r^2$=-0.254; p=N.S.). 

These results seem to indicate that both questionnaires represent a fair valuation of the different individual traits of the participants.

\subsection{Relation of extroversion to gaze and speech}\label{sec:extro}

The participants' average extroversion score was 111.77 ($\sigma$=22.86; min=61, max=160), which is, according to \cite{NEOPIR1998}, a neutral level of extroversion\footnote{According to the NEO-PIR, a participant obtaining a score bigger than 137 is considered extrovert, while one with a score below 80 is introvert.}. 

\begin{figure}
\centering
\includegraphics[width=0.99\hsize]{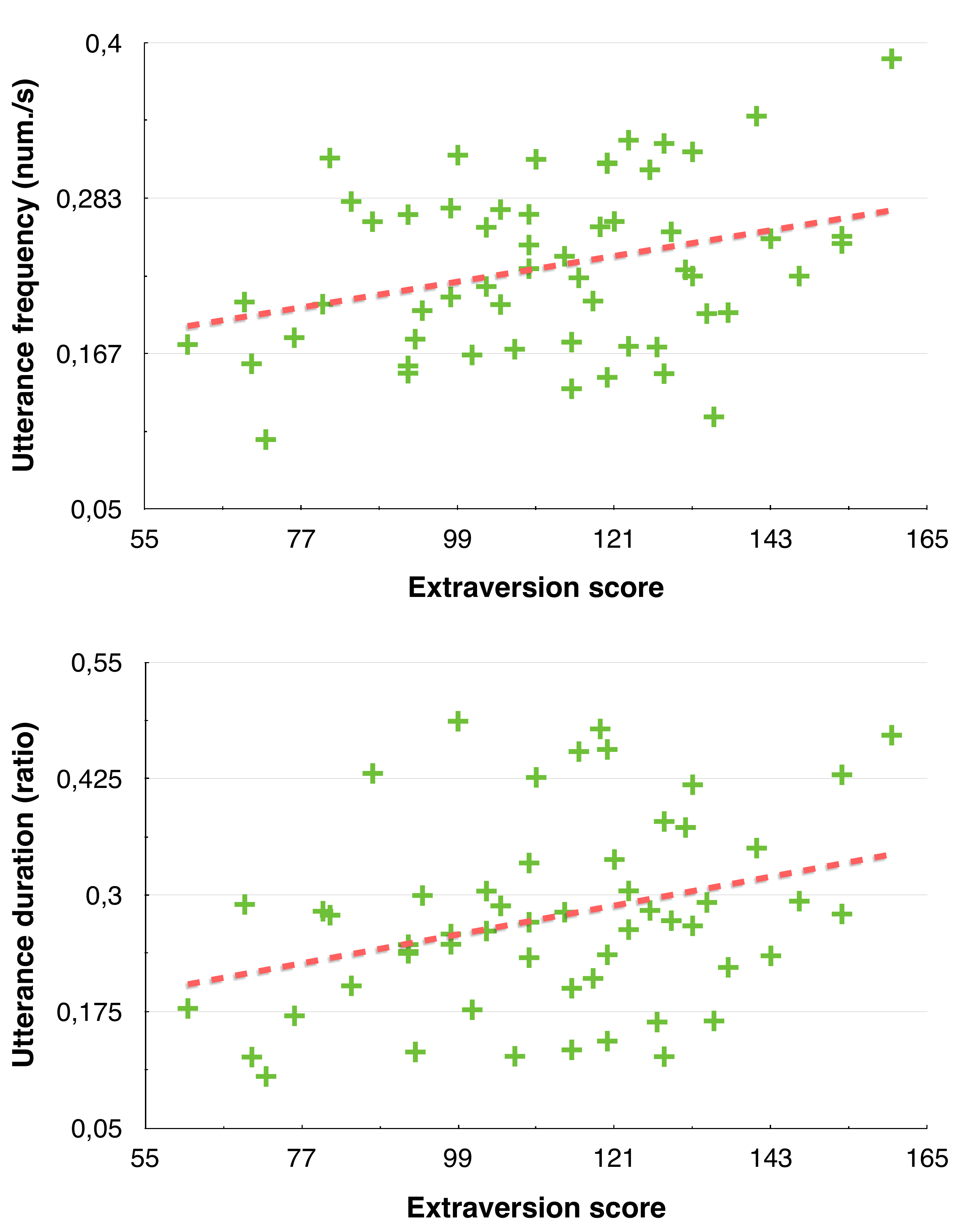}
\caption{Scatter graphs showing the frequency (number/seconds) and duration (normalized ratio) of utterances of the participants (N=56), in function of their extroversion score.}
\label{fig:extroversionutterance}
\end{figure}

Table~\ref{table:extroversion} reports the Pearson's correlation between the extroversion score of the participants and their gaze and utterance frequency and duration. 
The extroversion score is significantly and positively correlated to the frequency and duration of utterances (see Table~\ref{table:extroversion}). This can also be seen in the scatter graphs in Figure~\ref{fig:extroversionutterance}.
Conversely, the results indicate that extroversion does not influence the gaze signal, as there is no significant correlation between the personality trait and the gaze frequency or the duration of gaze.

To summarize, the more an individual is extrovert, the more he/she will tend to talk to the robot during an assembly task to provide instructions. 
On the contrary, an individual with a high score of extroversion will not look at the robot's face or hands more than individuals with lower scores.

Therefore, with reference to the research hypothesis expressed in Section~\ref{sec:hypotheses}, we confirm Hypothesis 1, and reject Hypothesis 2.

\begin{table*}
\centering
\begin{tabular}{|p{6cm}|p{4.5cm}|}
\hline
Variable & \textbf{Extroversion score}  \\
\hline
\hline
Gaze towards face frequency  & $r^2$= -0.13 ; p=0.927 (N.S.)  \\
Gaze towards face duration  & $r^2$= 0.098 ; p=0.471 (N.S.) \\
\hline
Gaze towards hands frequency  & $r^2$= 0.058 ; p=0.671 (N.S.)  \\
Gaze towards hands duration  & $r^2$= 0.215 ; p=0.875 (N.S.) \\
\hline
\textbf{Utterance frequency}  &	\textbf{$\mathbf{r^2}$= 0.318 ; p=0.017 ($\mathbf{<}$0.05)}  \\
\textbf{Utterance duration} &	\textbf{$\mathbf{r^2}$= 0.321 ; p=0.016 ($\mathbf{<}$0.05)}  \\
\hline
\end{tabular}
\caption{Correlation between the participants' extroversion score (computed by NEO-PI-R \cite{NEOPIR1998}) and their gaze and utterance frequency (number/s) and duration (normalized ratio) during the assembly task.}
\label{table:extroversion}
\end{table*}

\subsection{Relation of negative attitude towards robots to gaze and speech}\label{sec:nars}

The participants' average score for the negative attitude was 45.55 ($\sigma$=12.74; min=20, max=77), which is a neutral value for the attitude towards robots\footnote{According to the NARS, a score over 65 is a sign of negative attitude towards robots, while a score below 35 indicates a rather positive attitude towards robots.}. 

\begin{figure}
\centering
\includegraphics[width=0.99\hsize]{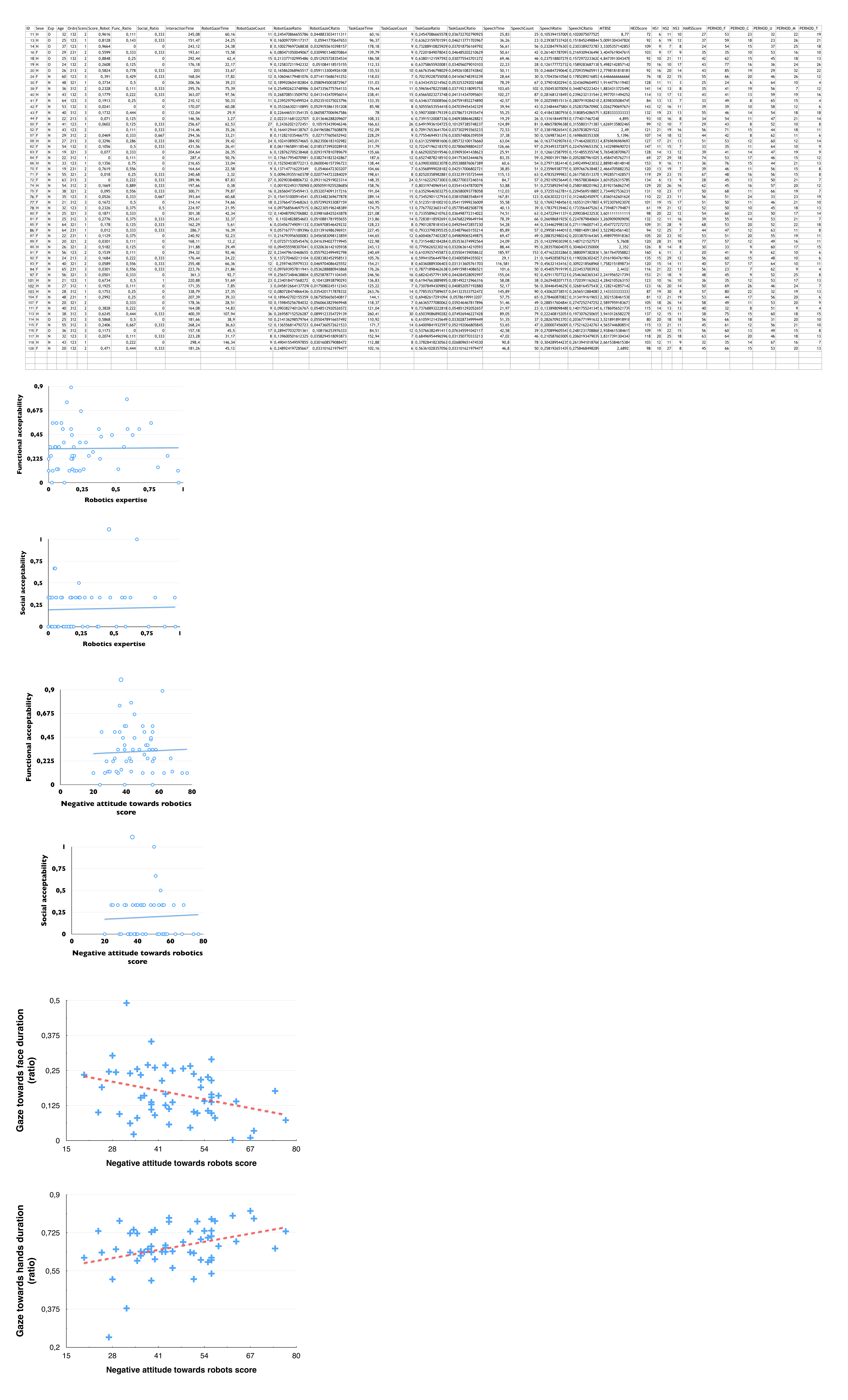}
\caption{Scatter graph showing the duration of gaze (normalized ratio) of the participants (N=56) towards the robot hands and face, in function of their NARS score. }
\label{fig:narsgaze}
\end{figure}

Table~\ref{table:nars} reports the Pearson's correlation between the NARS scores of the participants and their gaze and utterance frequency and duration. The results indicate that the negative attitude does not influence the verbal signal, as there is no significant correlation with the utterance frequency or duration. 
There is, however, a partial effect on the gaze signal. Precisely, the negative attitude is significantly and negatively related to the duration of gaze towards the robot's face, and positively related to the duration of gaze towards the robot's hands, as visible in Figure~\ref{fig:narsgaze}. 

To summarize, the more an individual has a negative attitude towards robots, the less he/she will tend to look at the robot's face during an assembly task, and the more he/she will tend to look at the robot's hands (area of physical contact). The gaze frequency, on the contrary, will not change in relation to different positive or negative attitudes. 
Nothing can be concluded regarding the verbal communication: an individual with a more negative attitude towards robots will not speak more or less to the robot than other individuals with a more positive attitude. 

Therefore, with reference to the research hypothesis expressed in Section~\ref{sec:hypotheses}, we reject Hypothesis 3
and partially confirm Hypothesis 4 and 5, since the NARS score relates to the gaze duration but not to the gaze frequency.

\begin{table*}
\centering
\begin{tabular}{|p{6cm}|p{8cm}|}
\hline
Variable & \textbf{Negative attitude towards robots score (NARS)} \\
\hline
\hline
Gaze towards face frequency  &  $r^2$= -0.174 ; p=0.201 (N.S.) \\
\textbf{Gaze towards face duration}  & \textbf{$\mathbf{r^2}$= -0.331 ; p=0.013 ($\mathbf{<}$0.05)} \\
\hline
Gaze towards hands frequency  &  $r^2$= -0.111 ; p=0.413 (N.S.) \\
\textbf{Gaze towards hands duration}  & \textbf{$\mathbf{r^2}$= 0.355 ; p=0.007 ($\mathbf{<}$0.05)} \\
\hline
Utterance frequency  &	$r^2$= -0.137 ; p=0.314 (N.S.) \\
Utterance duration  &	$r^2$= 0.033 ; p=0.807 (N.S.) \\
\hline
\end{tabular}
\caption{Correlation between the participants' negative attitude towards robots score (computed by NARS \cite{NARS2006}) and their gaze and utterance frequency (number/seconds) and duration (normalized ratio) during the assembly task.}
\label{table:nars}
\end{table*}

\begin{table*}
\centering
\begin{tabular}{|p{5.5cm}|p{2.9cm}|p{2.9cm}|p{2.9cm}|}
\hline
Variable & \textbf{NARS-S1} & \textbf{NARS-S2} & \textbf{NARS-S3} \\
\hline
\hline
Gaze towards face frequency  & $r^2$=-0.160; p=0.238 (N.S.)	 & $r^2$=-0.215; p=0.111 (N.S.)	& $r^2$=0.009; p=0.947 (N.S.) \\
\textbf{Gaze towards face duration}  & \textbf{$\mathbf{r^2}$=-0.311; p=0.020 ($\mathbf{<}$0.05)}	& \textbf{$\mathbf{r^2}$=-0.334; p=0.012 ($\mathbf{<}$0.05)}	& $r^2$=-0.120; p=0.377 (N.S.)\\
\hline
Gaze towards hands frequency  & $r^2$=-0.073; p=0.592	(N.S.) & $r^2$=-0.138; p=0.310	(N.S.) & $r^2$=-0.043; p=0.754 (N.S.)\\
\textbf{Gaze towards hands duration}  & \textbf{$\mathbf{r^2}$=0.381; p=0.004 ($\mathbf{<}$0.05)}	& \textbf{$\mathbf{r^2}$=0.334; p=0.012	($\mathbf{<}$0.05)} & $r^2$=0.094; p=0.491 (N.S.)\\
\hline
\textbf{Utterance frequency}  & $r^2$=0.018; p=0.895 (N.S.)	& $r^2$=-0.093; p=0.497	 (N.S.)& \textbf{$\mathbf{r^2}$=-0.323; p=0.015 ($\mathbf{<}$0.05)} \\
Utterance duration   &	$r^2$=0.172; p=0.203 (N.S.)	& $r^2$=0.058; p=0.673 (N.S.) &	$r^2$=-0.249; p=0.063 (N.S.) \\
\hline
\end{tabular}
\caption{Correlation between the scores of the NARS sub-scales (computed as in \cite{NARS2006}) of the participants and their gaze and utterance frequency (number/seconds) and duration (normalized ratio) during the assembly task.}
\label{table:narsspecific}
\end{table*}

As explained in Section~\ref{sec:material}, the NARS questionnaire is based on three sub-scales.
The participants' average scores of negative attitude towards interaction situations (S1), social influence of robots (S2) and emotions during interaction (S3) were respectively 15.18 ($\sigma$=5.83), 18.80 ($\sigma$=5.83) and 11.70 ($\sigma$=3.82), whereas the mean values of the three sub-scales were 24, 20 and 12. 
We performed a thorough investigation of the effect of each of the three sub-scales on gaze and utterances. 
For the gaze signal, we did not find any significant correlation between the sub-scales values and its frequency; however, we found a significant and negative correlation between the gaze duration and S1 ($r^2$=-0.311; \textbf{$p<0.05$}) and S2 ($r^2$=-0.334; \textbf{$p<0.05$}).
For the verbal signal, we did not find any significant correlation between the sub-scales values and the utterance duration, however we found a significant negative correlation between S3 and the utterance frequency ($r^2$=-0.323 ; \textbf{$p<0.05$}).

To summarize, the more people display a negative attitude \textit{in the interaction} (S1) with the robot and are concerned by the \textit{social aspect} (S2) of the interaction, the less they will look at the robot. Conversely, the more people are negative about the \textit{emotions during the interaction} (S3), the less they will talk to the robot.

\subsection{Post-experiment evaluation}

The post-experimental questionnaire for subjective evaluation does not have a score. 
It was designed by the experimenter to get a simple feedback on the user experience, retrieve the global impression and the personal evaluation of the participants on some aspects of the task.
Table \ref{table:postexperimentquestionnairescores} reports on the average score for each item in the questionnaire. We highlighted in bold the most significant questions, which have an average score that is close to the maximum (7) and minimum (1).

\section{Discussion}


As discussed in Section \ref{sec:background}, the literature in psychology highlights an effect of personality traits, particularly of the extroversion dimension, on the dynamics of speech and gaze. Likewise, a negative attitude towards robots will influence the time of the verbal response during interactions.
These results induced us to question the reliability of the metrics used for the estimation of the engagement in HRI, classically based on the dynamics of social signals, as their dynamics may be altered by individual factors that are not taken into account in the models of engagement.

In the following we discuss here the results on the correlations between two individual factors (extroversion and negative attitude towards robots) and the dynamics of speech and gaze observed during the human-robot assembly task. We argue about the implications of our study for the HRI community and consider the limits of our study.

\subsection{Extroversion \& social signals}
As detailed in Section~\ref{sec:extro}, we found that there is a positive and significant correlation between the extroversion score and the frequency and duration of utterances.
The more the individual is extrovert, the more often and longer he/she will tend to address the robot during the interaction. 
This result is consistent with observations of human-human interactions, showing that introverts tend to talk less than extroverts \cite{Scherer1981}. 
Conversely, we did not find a significant correlation between the extroversion and the gaze frequency or duration. This finding is partially contrary to what has been observed in \cite{Iizuka1992}, where the author found a relationship between the extroversion and the amount of time spent gazing while listening. However, the author also observed that the gaze duration was not related to extroversion when people were speaking. Since in our task, the participants were supposed to talk to the robot to explain the task, we can presume that this could be one possible cause of the non-effect of the extroversion on gaze duration. Furthermore, our assembly task induced the participants to focus their attention also on the robot hands, while we can presume that a different task will let people gaze at the robot face more frequently.
Another element that might explain this result is the lack of a proper joint attention system implemented on the robot for this experiment, particularly for mutual gaze: once the human touched the robot arms to start its kinesthetic demonstration, the robot was simply shifting its gaze from the human face to its own hands, and was not seeking eye-contact during the teaching phase.
However, since the participants were all in the same conditions, a correlation between gazing behavior and extroversion should have been detected.

\subsection{Negative attitude towards robots \& social signals} 
As presented in Section~\ref{sec:nars}, we found that the negative attitude towards robotics tends to be related to the time spent looking at the robot's face and the robot's hands during the interaction (Table \ref{table:nars}).

Overall, the participants were probably not apprehensive facing the interaction, while they were likely mildly concerned regarding the social and emotional aspect of the interaction.
With a deeper look at the NARS sub-scales (Table \ref{table:narsspecific}), we found that the more one has a negative attitude towards the interaction situation (S1) and the social influence of the robot (S2), the shorter it will look at the robot face. 
These results may indicate that people will look less at the robot as symptom of their aversion towards the robot as social agent, or because of their anxiety in interacting with it. 
This is consistent with the duration of gaze directed towards the robot's hands: it makes sense that the more one has a negative attitude towards interacting with a robot, the more he/she will spend time looking at the robot's hands in a task where there is physical interaction with the robot occurring at the hands level.
Interestingly, these dimensions (S1 and S2) do not seem to have influence on the speech production. 
Conversely, people concerned with emotional robots (S3) will tend to have less verbal exchange with the robot. 

We found significant correlations for the gaze duration, but not for the gaze frequency: this result could be slightly biased by the lack of mutual gaze exhibited by the robot.
We expected that an individual with positive attitude would look more at the face trying to engage and get the robot's attention, while an individual with negative attitude would have the tendency to avert his/her look towards the robot face.
However, the lack of a joint attention mechanism can explain the low number of gazes towards the robot face (12.13 $\pm$ 6.57) and the fact that they do not seem to be correlated with the negative attitude.

We did not find any significant correlation with the verbal signal.
Our results seem to contradict those of \cite{Nomura2008}, that brought evidence that a negative attitude towards robots had repercussions on the timing of the verbal response. However, in their study the authors were focusing on reaction times to robot's stimuli, not on the frequency or duration of utterances.
Looking at the NARS sub-scale, we found a significant correlation between the negative attitude towards emotions (S3) and the utterance frequency. This result is in line with \cite{NARS2006}, where the authors highlight the stronger negative attitude towards emotions (S3) for individuals dealing with small-sized humanoids robots, which corresponds to the case of our robot iCub.

Overall, we expected the negative attitude to have a stronger influence on the amount of verbal and non-verbal signals exchanged during the interaction. 
We expected that the physical contact with the robot and the close interaction would particularly highlight the effect of the negative attitude. 
We speculate that this result could be influenced by a social desirability bias: the participants maybe tried to perform better as subjects in the study, eventually behaving in a forced way. The positive evaluation that we retrieved from the post-experimental questionnaire (Table \ref{table:postexperimentquestionnairescores}) could also be partially related to that.

As we found few studies dealing with attitude towards robots and social signals, this part of our work may be considered as exploratory.

\subsection{Subjective impressions}

Overall, the subjective evaluations and the feedback from the interviews encourage us to think that the interaction with the robot was pleasant and the participants were spontaneous in their behavior.
With reference to the subjective evaluations scores in Table \ref{table:postexperimentquestionnairescores}, the participants evaluated positively the experiment with the robot and the robot itself. We highlighted in bold the questions where the average score is close to the maximum (7) or minimum (1) score: this provides a rough indication of the tendencies of the participants.
They found the task quite interesting and easy to do, and they also had a positive impression of the robot.  
Interestingly, they were not afraid to touch or interact physically with the robot (e.g., not worried to touch the robot, not afraid to touch the hands). Also the robot was not looking dangerous to their eyes.
Considering that the experiment was their first live interaction with the robot, this score was quite surprising for us: we expected the novice/naive people to report some anxiety in front of the robot. However, when we interrogated the participants about this, most of them said that the safety demonstration reassured them about the fact that it was possible to touch the robot without problems; others said that the robot size and child-like appearance made them suppose that it was safe to touch it as the robot ``won't hurt''.
We asked to the participants if they thought or had the impression that the robot was operated by someone else: all the participants denied this possibility. Almost all the participants asked us if the robot had learned correctly what they had been teaching.

\subsection{Implications for automatic personality assessment}


Social robots should be able to adapt their behaviour taking into account the unique personality of their interacting partners \cite{Anzalone2012profile}. To this end, they need to learn a model of their behaviour, that can be built using multimodal features extracted during online interaction, physical features, social context, psychological traits of the partner such as personality or attitudes etc.
Currently, a crucial challenge in HRI is the automated online estimation of these individual factors: for personality traits, in the \emph{personality computing} literature this is called Automatic Personality Recognition, which aims at ``inferring self-assessed personalities from machine detectable distal cues'' (see \cite{Vinciarelli14} for an exhaustive review).  
Since personality and individual traits influence the production of verbal and non-verbal signals, it is important to gain more quantitative knowledge on their relations to be able to produce predictive models that can be used to improve the HRI experience.
For example, Tapus et al. \cite{Tapus08,Tapus08b} showed that an adaptive robot matching the personality of the patient is beneficial for assisted therapy, and that extrovert/introvert people prefer to interact with robots that exhibit extrovert/introvert personality features \cite{Aly2013personality}. 

Thanks to the findings of our work, we now have a quantitative indicator for estimating the extroversion of a human interacting with a robot in a collaborative assembly task, by looking at the his/her speech dynamics. At the same time, we can derive a simple linear model for estimating the NARS based on the duration of gaze towards the robot face. 

We are extending these findings to the other experiments of the EDHHI project, for example we already showed that it is possible to predict extroversion from non-verbal features during a thin slice of simple face-to-face interaction \cite{ICSR2015}.

\subsection{Implications for measuring engagement}

Our goal in this paper is not to measure the engagement of a particular HRI situation, but to provide quantitative evidence that the computational models of engagement should take into account individual factors. Such models are commonly based on the dynamics of signals such as gaze and speech \cite{Anzalone2015engagement,sidner2004,rich2010recognizing}.
The engagement metrics may be biased by individual factors such as extroversion and negative attitude towards robots, factors that we have not met in the engagement literature for an assembly task such as the one presented in this paper.
Our results indicate that extroversion and negative attitude towards robots are related to the temporal dynamics of gaze and speech during a human-robot interaction.
If the engagement depends on the frequency or rhythm of such social signals \cite{rich2010recognizing}, then an introvert individual or one with negative attitude towards robots will be considered as less engaged than an extrovert or one with a positive attitude, simply because the first is more likely to produce less signals (gaze or utterances) than the second.
The design of robust models of engagement should therefore take these individual factors into account.

We further notice that the models for evaluating the engagement refer mostly to dyadic tasks without physical interaction. For tasks such as the one of this paper, the cooperative assembly may induce the people to gaze more at the hands and at the objects than in other tasks where there is no co-manipulation. 
Therefore, there is a potential risk that the estimated engagement of the HRI may be partially biased by the ``task engagement''. We will perform the study with other tasks to verify, because the current results are not sufficient to provide conclusions on this matter.
This problem actually highlights a weakness of the models used for the evaluation of the engagement which are uniquely based on the dynamics of social signals.

\subsection{Implications for human-robot physical interaction}

The underlying idea in our work is that by studying the factors that influence the production of social signals together with the exchanged forces, one can improve the design of robot controllers during physical interaction and, for example, achieve better performances during cooperative tasks.

More and more people are going to interact physically with robots, for a variety of tasks: from co-working in manufacturing, to personal assistance at home.
This requires for the robot the ability to control precisely the interaction forces, but also to be able to interact in a ``social'' way, adapting to the reaction of each individual, so that people can trust the robot, accept it and be engaged interacting with it.

Together with the contact forces, it is therefore necessary to study the other verbal and non-verbal signals that are exchanged during the physical interaction, such as gaze, prosody, gestures, etc.
All these signals can be used to study the comfort and the engagement of the people interacting with the robot, providing the necessary feedback for the robot to adapt its action.
Researchers studying cooperative tasks usually focus on sequencing and patterns of cooperation \cite{Wilcox2012}, adaptation of roles and physical forces \cite{Stefanov2009}, while the social signals emitted during such tasks are not fully explored.
Conversely, the dynamics of social signals, such as gaze and speech, is mostly studied during tasks that do not involve a direct physical interaction, such as dialogues and games \cite{Boucenna2014,Anzalone2015engagement,Castellano2009}.

In this paper, we provide some evidence about the dynamics of speech and gaze during a cooperative assembly task with physical interaction. To the best of our knowledge, this is the first work analyzing social signals during a cooperative assembly task with a humanoid robot.

\subsection{Limits of the study}

The present study brings significant new results to the field of human-robot interaction and engagement. However, we discuss the limitations of our study.

\subsubsection{Ordinary people}

In our study, participants interacting with the robot are not experts ``robot-users''. Our findings could change if we considered people with different levels of exposure to robotics and technology and expertise with iCub or other robots. 
Our intuition is that the prior exposure to robotics is likely to appear in the dynamics of verbal and non-verbal signals.
This question is currently under investigation.

\subsubsection{HHI vs HRI}

It would have been interesting to have a control study about human-human physical interaction for the same assembly task. This kind of study would enable to compare if the dynamics of the social signals emitted by the human change when interacting with a human or with a robot during a physical collaboration, in function of the individual factors of the human. 
However, the same experiment done by two humans would have been too different in our view, and not only because the engagement of human-human and human-robot interaction is different.
In our experiment, iCub is a child-like robot, and the task is very simple: it would have been difficult to make it engaging for two adults, and would have made sense to do it with an adult and a child. However, the child should have been constrained to be basically not too reactive. 
We actually did, in a preliminary investigation, record the assembly task performed by a father and his child, two sisters (one older than the other) and two adults. Despite our instructions to the children, we found very difficult to reproduce the experiment with similar conditions to the ones of the HRI experiment. 
For example, 
it was difficult for one to not to react to the action of the other:
we observed anticipatory gaze, joint gaze, anticipatory movements of the arms before and during the kinesthetic teaching, etc. These mechanisms were not implemented in our robot. Empathy, personality traits and other factors linked to the human partner acting as the robot should also have been taken into account.

\subsubsection{Generalization}

In this study, we focused on an assembly task requiring physical interaction. The situation addressed in this study is extremely relevant to the robotics community and particularly to those studying collaborative robots and robotic co-workers. It is difficult to predict whether our results can be generalized to other tasks. This question is currently under investigation.

\subsubsection{Human-like and child-like appearance}

Another limit of our study is given by the human-like appearance of the robot, which may influence the production of social signals. This question was equally raised in other studies with human-like robots, for example by Huang and Thomaz with the Meka robot \cite{Huang2011}.
As we already remarked in our previous studies with iCub \cite{ivaldi2014frontiers}, the anthropomorphic appearance may induce a biased perception of the robot and ultimately influence the dynamics of speech and gaze, especially the one directed towards the robot face. 
However, differently from the previous study, before the experiments we told the participant that the robot had a limited knowledge and they had to teach the robot how to build the object. As their expectations about the robot intelligence were lower, their subjective evaluation of the robot resulted to be globally more positive than the one of the previous experiment (see Table \ref{table:postexperimentquestionnairescores}). 
The type of task could also play a role: here the participant had a very close interaction with the robot, and had to use the hands of the robot for building an object. The task implies manipulation skills and cognitive skills that are usually attributed to humans and intelligent agents. For example, learning to ``align'' the cylinders means learning the proper arm movements but also understanding the concept of ``to align'' and ``to assemble an object made by two parts.'' Some participants were so engaged with the robot and the task that spent time to make sure that the robot could learn these concepts, showing the assembly gesture before engaging the kinesthetic teaching, and showing the final object explaining the result of the action after the kinesthetic teaching (see Figure \ref{fig:shotsfinished}).
It is also possible that the child-like appearance of the robot facilitated the emergence of these behaviors. However, we did not consider in our study the attitude towards children or having children as possible individual factors: this is a limitation of the study.

Would the results be different with another type of robot? For example a collaborative industrial robot without an anthropomorphic head? We are currently investigating this question.

\section{Conclusions}

In this paper we reported on the influence of extroversion and attitude towards robots on the temporal dynamics of social signals (i.e., gaze toward the robot's face and speech), during a human-robot interaction task, where a human must physically cooperate with a robot to assemble an object. 

We conducted the experiments with the humanoid robot iCub and N=56 adult participants. 
We focused on extroversion and negative attitude towards robots, and reported on their effect on gaze and speech.

We found that the more people are extrovert, the more they tend to talk and longer to the robot. We also found that the more people have a negative attitude towards robots, the less they tend to look at the robot's face. 

The assembly task entailed a physical contact between the human and the robot: we found that the more people have a negative attitude towards robots, the more they look at the area where the physical contacts occurred and the assembly task was executed  (in this case, the robot's hands). 

Our results provide evidence that among the metrics classically used in human-robot interaction to estimate engagement~\cite{rich2010recognizing}, one should also take into account inter-individual factors such as extroversion and attitude towards robots, because these individual factors influence the dynamics of social signals, hence the dynamics of the interaction. 
Furthermore, we highlight a potential risk for the classical models of engagement, that do not provide a solution to the problem of decoupling the engagement towards the robot and the engagement towards the task. These two are not easily discernible from the study of social signals, for many cooperative tasks.

To summarize, we propose an original way to deal with engagement and social signals during HRI: with a more comprehensive and multidisciplinary approach, we explicitly consider the exchanged social signals and the individual factors that may influence the production of such signals. Particularly, we do not only consider the personality traits of the humans, but also their attitudes towards robots that may be critical for their behavior during the interaction with a robot. 
  
The influence of personality traits on social signals should be taken into account if we wish to build robots capable of automatically estimating the engagement of the human partner - in tasks with or without physical interaction.
Of course, other dimensions should be investigated, for instance individual traits (e.g., other personality dimensions from the Big-Five~\cite{BIGFIVE}, such as openness or neuroticism), social attitudes or environmental and contextual factors. 
Recent studies show that it is possible to retrieve personality traits online from audio or video streams~\cite{mohammadi2012automatic}. 
It will be therefore feasible to pair the personality estimation with the social signals analysis, to provide better models of human engagement. Such models will be critical to adapt the robot's behavior to the single individual reaction.

Our insights can play an important role for letting the robot adapt its behavior to the human response. For example, to re-engage the dis-engaged partner into a cooperation by means of relevant social signals or physical actions.

\section{Questionnaire for negative attitude towards robots (NARS)}\label{appendix:nars}

See Table \ref{table:nars} for the questions in English and French.

\begin{table*}
\textbf{Negative Attitude Towards Robots Questionnaire (NARS)}

\begin{tabular}{|c|p{7cm}|p{7cm}|c|}
\hline
N. & Questionnaire Item in English & Questionnaire Item in French & Subscale \\
 \hline
 \hline
1& I would feel uneasy if robots really had emotions. & Je me sentirais mal à l'aise si les robots avaient réellement des émotions. & S2 \\
2& Something bad might happen if robots developed into living
beings. & Quelque chose de mauvais pourrait se produire si les robots devenaient des êtres vivants. & S2 \\ 
3& I would feel relaxed talking with robots. & Je serais détendu(e) si je parlais avec des robots. & S3* \\
4& I would feel uneasy if I was given a job where I had to use robots. & Je me sentirais mal à l'aise dans un travail où je devrais utiliser des robots. & S1 \\
5& If robots had emotions, I would be able to make friends with them. & Si les robots avaient des émotions, je serai capable de devenir ami(e) avec eux. & S3 \\
6& I feel comforted being with robots that have emotions. & Je me sens réconforté(e) par le fait d’être avec des robots qui ont des émotions. & S3* \\
7& The word “robot” means nothing to me. & Le mot ‘‘robot’’ ne signifie rien pour moi. & S1 \\
8& I would feel nervous operating a robot in front of other people. & Je me sentirais nerveux/nerveuse de manœuvrer un robot devant d'autres personnes. & S1 \\
9& I would hate the idea that robots or artificial intelligences were
making judgments about things. & Je détesterais que les robots ou les intelligences artificielles fassent des jugements sur des choses. & S1 \\
10& I would feel very nervous just standing in front of a robot. & Le simple fait de me tenir face à un robot me rendrait très nerveux/nerveuse. & S1 \\
11& I feel that if I depend on robots too much, something bad might
happen. & Je pense que si je dépendais trop fortement des robots, quelque chose de mauvais pourrait arriver. & S2 \\
12& I would feel paranoid talking with a robot. & Je me sentirais paranoïaque de parler avec un robot. & S1 \\
13& I am concerned that robots would be a bad influence on children. & Je suis préoccupé(e) par le fait que les robots puissent avoir une mauvaise influence sur les enfants. & S2 \\
14& I feel that in the future society will be dominated by robots. & Je pense que dans le futur la société sera dominée par les robots. & S2 \\
\hline
\end{tabular}
* = reverse item

\caption{NARS questionnaire for evaluating the negative attitude towards robots. The order of the questions follows the original questionnaire, proposed by Nomura et al. in \cite{Nomura2006nars}. The second column reports the original questions in English. The third column reports our double translation of the questions in French.}
\label{table:nars}
\end{table*}

\section{Questionnaire for post-experimental evaluation of the assembly task}\label{appendix:postexpquestionnaire}

See Table \ref{table:postexperimentquestionnaire} for the questions in English and French.

\begin{table*}
\textbf{Post-experimental questionnaire for evaluation of the human-humanoid collaborative tasks with physical interaction}

\begin{tabular}{|c|p{7.5cm}|p{7.5cm}|}
\hline
N. & Questionnaire Item in English & Questionnaire Item in French  \\
 \hline
 \hline
1 & The assembly task was easy to do. & La tâche de constructions était facile à faire.\\
2 & The assembly task was interesting to do.& La tâche de construction était interessante à faire.	\\
3 & I was worried to must touch the robot to assembly the objects with it. & J'etais inquiet(e) de devoir toucher le robot pour construire les choses avec lui.\\
4 & During the assembly, I would have preferred that the robot tells me what it thinks, if it understands well. & Pendant la construction, j'aurais préfèré que le robot m'informe de ce qu'il pense, s'il comprend bien.\\
5 & I was afraid to touch the hands of the robot. & J'avais peur de toucher les mains du robot.\\
6 & I was afraid to damage the robot. &	J'avais peur d'abimer le robot.	\\
7 & The robot was nice. & Le robot était sympathique.\\
8 & The robot understood what I explained to it.&	Le robot a compris ce que je lui ai expliqué. \\
9 & The robot answers to questions too slowly. & Le robot réponds aux questions trop lentement. \\
10 & The robot moves its head too slowly. & Le robot bouge la tête trop lentement.\\
11 & The robot moves its arms too slowly. & Le robot bouge les bras trop lentement.\\
12 & The robot should be more reactive.& Le robot devrait être plus réactif.\\
13 & The facial expressions of the robot trouble me.& Les expressions faciales du robot me gênent.\\
14 & The voice of the robot is pleasant.& La voix du robot est agreable.\\
15 & The robot is not threatening.& Le robot n'est pas menacant.\\
16 & The robot does not look dangerous.& Le robot ne semble pas dangereux.\\
17 & Someday I could work with this robot to build something of interest& Un jour, je pourrais travailler avec this robot pour construire quelque chose d'interessant\\
18&  Someday I could work with a robot to build something of interest& Un jour, je pourrais travailler avec un robot pour construire quelque chose d'interessant\\
\hline
\end{tabular}
\caption{Post-experimental questionnaire for evaluating the perception and interaction with the iCub in the assembly task of this work. The third column reports the original questions in French (the participants were all native French speakers). The second column reports our double translation of the questions in English.}
\label{table:postexperimentquestionnaire}
\end{table*}

\section{Software for operating the robot}\label{appendix:GUI}

The WoZ GUI was organized in several tabs, each dedicated to a specific task, such as controlling the robot movements (gaze, hands movements, posture), its speech, its face expressions etc.
The GUI events are elaborated by the actionServer module and others developed by the authors in previous studies \cite{ivaldi2014frontiers,Ivaldi2014tamd}. All the developed software is open source\footnote{See download instructions at \url{http://eris.liralab.it/wiki/UPMC_iCub_project/MACSi_Software}}.

\begin{figure*}
\centering
{\large \textbf{\textsf{A}}} \includegraphics[height=7cm]{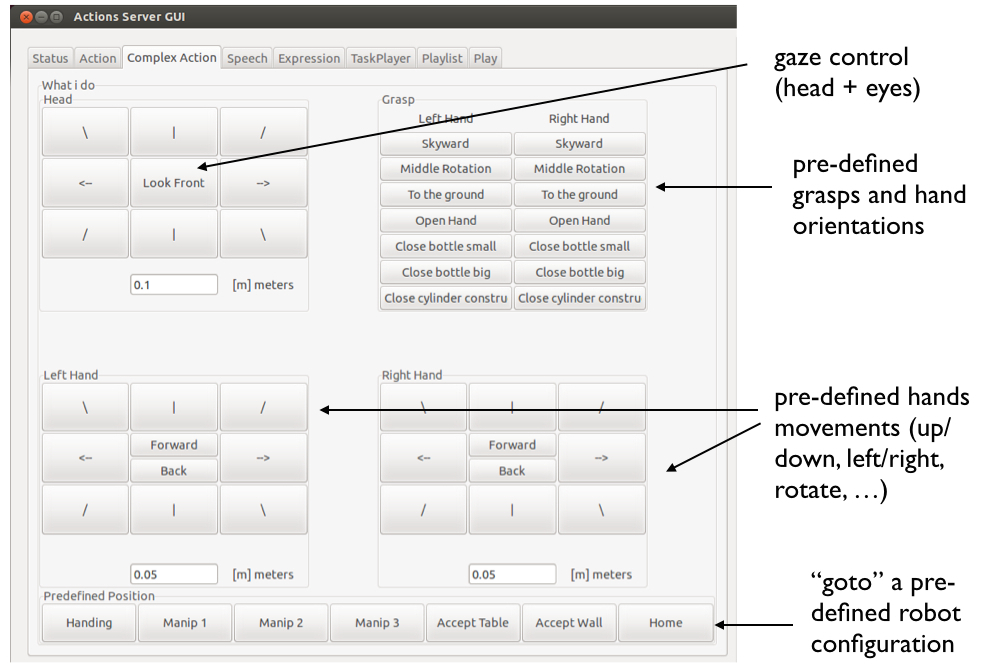} \hspace{0.5cm}
{\large \textbf{\textsf{B}}} \includegraphics[height=2.5cm]{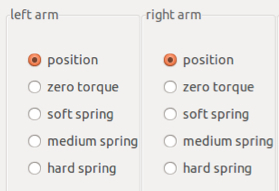}
\caption{WoZ GUI. \textbf{\textsf{A}}: the tab dedicated to the quick control of gaze, grasps and hands movements in the Cartesian space. The buttons sends pre-defined commands to the \textit{actionsServer} module, developed in \cite{Ivaldi2014tamd}. The buttons of the bottom row allows the operator to bring the robot in pre-defined postures (whole-body joint configurations): they were pre-programmed so as to simplify the control of the iCub during the experiments, in case the operator had to ``bring it back'' to a pre-defined configuration that could simplify the interaction for the participants. They were useful also for prototyping and testing of the experiments. \textbf{\textsf{B}}: part of the GUI dedicated to switching the control mode of the arms -- position, zero-torque, then impedance control with low, medium and high stiffness.}
\label{fig:guiactions}
\end{figure*}

\begin{figure*}
\centering
{\large \textbf{\textsf{A}}} \includegraphics[height=6cm]{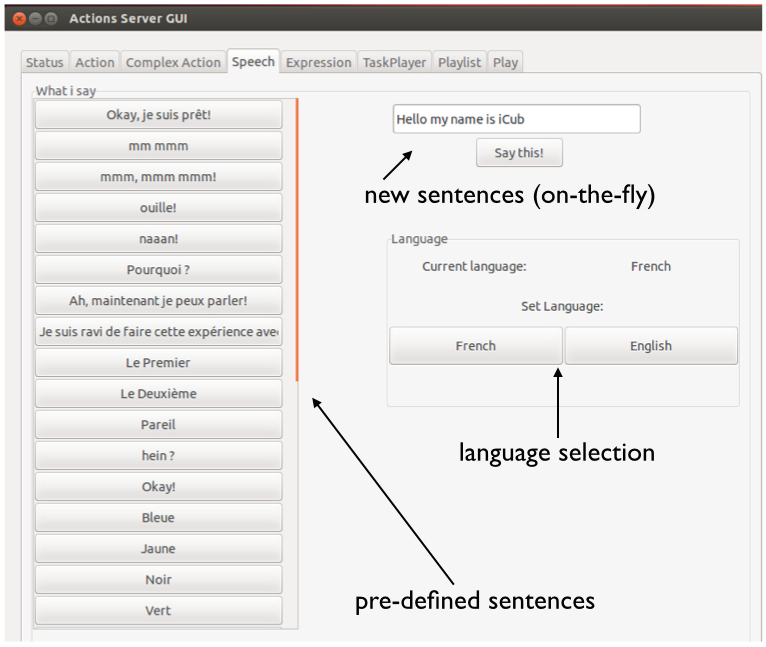} \hspace{0.5cm}
{\large \textbf{\textsf{B}}}  \includegraphics[height=6cm]{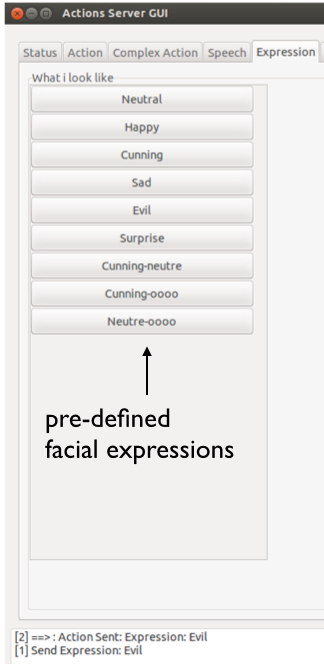}
\caption{WoZ GUI. \textbf{\textsf{A}}: the tab related to the robot's speech. The operator can choose between a list of pre-defined sentences and expressions, or he can type a new sentence on-the-fly: this is done to be able to quickly formulate an answer to an unexpected request of the participant. The operator can switch between french and english speech (at the moment, the only two supported languages), even if in the experiments of this paper of course the robot was always speaking french. \textbf{\textsf{B}}: the tab related to facial expressions. The list of facial expression along with their specific realization on the iCub face (the combination of the activation of the LEDs in eyelids and mouth) is loaded from a configuration file.}
\label{fig:guispeech}
\end{figure*}

Figure \ref{fig:guiactions}-A shows the tab related to the control of head gaze and hands movements. It is designed to control the gaze direction in the Cartesian space, with relative movements with respect to the fixation position (joints at zero degrees in both eyes and neck). The hands can be quickly controlled by a list of available pre-defined grasps, plus primitives for rotating the palm orientation (towards the ground, skywards, facing each other). It is also possible to control the hand position and orientation in the Cartesian space, providing relative movements with respect to the current position with respect to the Cartesian base frame of the robot (the origin located at the base of the torso, with x-axis pointing backward, y-axis pointing towards the right side of the robot and z-axis pointing towards the robot head). Some buttons allow the operator to control the whole posture of the robot and bring it back to pre-defined configurations. 
Figure \ref{fig:guiactions}-B shows the part of the GUI dedicated to switching the control mode of the arms: position, zero-torque, then impedance with high, medium and low stiffness. The default values of the module \textit{demoForceControl}\footnote{\url{https://github.com/robotology/icub-basic-demos/tree/master/demoForceControl}} for stiffness and damping were used. During the experiments, the arms were controlled in the ``medium compliance'' impedance mode, which allows the robot to exhibit a good compliance in case of unexpected contacts with the human participant. When the participant had grabbed the robot arms to start the teaching movement, the operator switched the control to zero-torque, which made the arms move under the effect of the human guidance. 
Figure \ref{fig:guispeech}-A shows the tab related to the robot's speech. It is designed to quickly choose choose one among a list of pre-defined sentences and expressions, in one of the supported languages (currently French or English). It is also possible to generate new sentences, that can be typed on-the-fly by the operator: this is done to allow the operator to quickly formulate an answer to an unexpected request of the participant. The operator can switch between the supported languages, but of course in the experiments of this paper the robot was always speaking French (as all the participants were native french speakers). The text-to-speech in English is generated by the \texttt{festival} library, while in French by the \texttt{Pico} library. 
Figure \ref{fig:guispeech}-B shows the tab related to facial expressions. The list of facial expressions along with their specific realization on the iCub face (the combination of the activation of the LEDs in eyelids and mouth) is loaded from a configuration file that was designed by the experimenter. 

\begin{acknowledgements}
The authors wish to thank Charles Ballarini for his contribution in software and experiments, Salvatore Anzalone and Ilaria Gaudiello for their contribution to the design of the experimental protocol.
\end{acknowledgements}

\bibliographystyle{spmpsci}      
\bibliography{edhhi}   

\end{document}